\documentclass[twoside,11pt]{article}
\usepackage[preprint]{jmlr2e}
\usepackage{amsmath,mathtools,booktabs,array,graphicx,microtype}
\usepackage{tikz}
\definecolor{vermil}{HTML}{D55E00}
\usepackage[section]{placeins}
\usepackage{caption}
\hypersetup{pdfborder={0 0 0},pdftitle={Interaction Retention and Output-Likelihood Training in Parallel Sampling},pdfauthor={Da Li, Bo Jin, Xiaopeng Wei, Hui Wang}}
\renewenvironment{proof}[1][Proof]{\par\noindent{\bfseries #1\ }}{\hfill\BlackBox\\[2mm]}
\providecommand{\qed}{\hfill\BlackBox}
\newtheorem{assumption}{Assumption}
\newcommand{\E}{\mathbb E}
\newcommand{\Prb}{\mathbb P}
\newcommand{\KL}{D_{\mathrm{KL}}}
\newcommand{\TC}{\mathrm{TC}}
\newcommand{\calC}{\mathcal C}
\newcommand{\calH}{\mathcal H}
\newcommand{\calL}{\mathcal L}
\newcommand{\norm}[1]{\left\lVert #1\right\rVert}
\newcommand{\inner}[2]{\left\langle #1,#2\right\rangle}
\newcommand{\ind}{\mathbf 1}

\ShortHeadings{Interaction Retention in Parallel Sampling}{Li, Jin, Wei and Wang}
\firstpageno{1}

\begin{document}

\title{Interaction Retention and Output-Likelihood\\ Training in Parallel Sampling}

\author{\name Da Li \email dli@mail.dlut.edu.cn \\
       \name Bo Jin\thanks{Co-corresponding authors.} \email jinbo@dlut.edu.cn \\
       \name Xiaopeng Wei\footnotemark[1] \email xpwei@dlut.edu.cn \\
       \addr School of Computer Science and Technology\\
       Dalian University of Technology\\
       Dalian, Liaoning, China
       \AND
       \name Hui Wang\footnotemark[1] \email hwang1606@dicp.ac.cn \\
       \addr Dalian Institute of Chemical Physics\\
       Chinese Academy of Sciences\\
       Dalian, Liaoning, China}

\maketitle

\begin{abstract}
Parallel samplers such as masked diffusion language models generate several coordinates per step from conditional marginals and randomize which coordinates are generated together, so even exact conditionals return a distorted distribution. We give an exact local account of this distortion. Around any full-support product law, the Gram operator of the sampler's response is diagonal in the interaction decomposition, and its eigenvalues are retention probabilities: the probability that an interaction has a unique latest-revealed coordinate. This retention spectrum determines the exact frontier between output accuracy and conditional calibration, its regularization path, and the population response of finite-mixture likelihood training. For pairwise interactions it yields a sharp dichotomy. Under homogeneous retention, output-likelihood training is equivalent to a single context gain; heterogeneous retention opens correction directions beyond any scalar, with an advantage bounded by an explicit heterogeneity gap. On binarized digits and MNIST, output-likelihood training within a heterogeneous schedule library lowers calibrated test negative log-likelihood in all twelve paired comparisons, and the gain survives pair-specific recalibration and a stronger path optimizer. For a frozen 169.6-million-parameter masked diffusion language model, serially normalized context gains are 1.070 and 1.056 at 8 and 16 rounds, against the predicted values 1.134 and 1.058.
\end{abstract}

\begin{keywords}
parallel sampling, masked diffusion models, randomized schedules, interaction decomposition, mixture likelihood
\end{keywords}

\section{Introduction}
Parallel generation is a central route to fast sampling from high-dimensional discrete models. Iterative parallel decoders for translation and images \citep{nat,maskpredict,maskgit} and masked diffusion models \citep{d3pm,sedd,mdlm,md4,llada} reveal several coordinates per step, drawing each from its conditional marginal given the coordinates generated so far. This trades serial computation for dependence error: coordinates drawn in the same step are conditionally independent by construction, so the returned distribution differs from the target even when every conditional marginal is exact. Large-scale evaluations attribute much of the remaining gap between masked diffusion and autoregressive language models to this loss of dependence \citep{zhong}. Practical samplers also randomize which coordinates are revealed together. Each sample executes a single schedule, but the returned distribution averages over the random choice of schedule. Understanding this averaged distribution matters both for choosing schedules and for training the conditional predictor that they query.

Two lines of work describe the error of parallel sampling. Information-theoretic analyses bound the error of a schedule through the mutual information, total correlation or information profile of the target and use these bounds to design schedules \citep{licai,chen,lavenant,zhaocai,wainwright}. For deterministic value-adaptive rules, the conditional total correlation gives an exact accounting of the dependence discarded along the reveal tree \citep{wen}. After mixing over schedules, these per-path losses no longer determine the output error, and the difference can be large \citep{li,wu}. What has been missing is a description of what a schedule mixture does to each individual dependency of the target, and hence of what a learned predictor must change to undo it.

In this paper we show that one spectrum answers both questions. Decompose a perturbation of a product law into orthogonal interactions $f_A$. An interaction can enter a scalar prediction only when all but one of its coordinates have been revealed, so a schedule transmits $f_A$ exactly when the interaction has a unique latest-revealed coordinate. Its \emph{retention} $c_A$ is the probability of this event (Figure~\ref{fig:retention}). Our main result, Theorem~\ref{thm:main}, shows that these probabilities are the spectrum of the sampler's response: the Gram operator of the sampling map equals $\sum_Ac_AP_A$. Exact conditionals therefore attenuate every interaction to $c_Af_A$, and restoring the missing $(1-c_A)f_A$ costs exactly $(1-c_A)^2\norm{f_A}^2/c_A$ in conditional calibration. A single spectral formula describes the entire tradeoff between output accuracy and calibration, and whenever every interaction present in the target has positive retention, shared kernels can reduce the output divergence to fourth order in the perturbation size.

The spectrum also explains what likelihood training does. Accurate conditional predictions and an accurate final distribution are different objectives for the same sampler. Training on the mean path likelihood targets the first; training on the likelihood of the final mixture, or on its $M$-sample importance-weighted bound \citep{iwae}, targets the second. Finite-mixture training moves along the same output--calibration frontier: the $M$-sample objective plus $\lambda$ times the path likelihood selects the frontier point of calibration weight $(M\lambda+1)/(M-1)$ (Theorem~\ref{cor:finiteM}). For pairwise interactions this leads to a sharp dichotomy (Corollary~\ref{cor:postscalar}). When all interactions share one retention, everything that output-likelihood training changes is reproduced by one post-hoc \emph{context gain}, a common multiplier on the contextual part of every logit, whose value under uniform batching is predicted without fitting to be $1/c_2$, the inverse of the pair retention (Corollary~\ref{cor:context-gain}). When positive retentions differ, output training reaches correction directions that no scalar reaches, and its advantage over the best scalar is bounded by a heterogeneity gap $H$, attained at the output-likelihood optimum \eqref{eq:heterogeneity-limit}. A full-support three-spin example certifies the strict advantage at finite coupling.

\paragraph{Contributions.}
\begin{enumerate}
\item \textbf{The retention spectrum.} We identify the response Gram operator of a randomized parallel sampler as $L_\pi L_\pi^*=\sum_Ac_AP_A$ and derive the exact output--calibration frontier (Theorem~\ref{thm:main}), its closed-form regularization path (Corollary~\ref{cor:regularized}) and an explicit finite-radius guarantee (Proposition~\ref{prop:finite}).
\item \textbf{Training response and a sharp dichotomy.} We characterize the population optimum of finite-mixture likelihood training (Theorem~\ref{cor:finiteM}), show that homogeneous retention reduces it to a scalar context gain while heterogeneous retention yields a strict advantage bounded by $H$ (Corollary~\ref{cor:postscalar} and \eqref{eq:heterogeneity-limit}), and certify the advantage on a full-support finite system.
\item \textbf{Consequences for schedule design.} The spectrum predicts the context gain for uniform batching without fitting, and gives the exact expected residual of finite random schedule banks \eqref{eq:finite-bank-residual}, the exact leading error of the best deterministic adaptive sampler on Ising models, a convex criterion for schedule weights and an interaction correction estimated from samples alone (Appendix~\ref{app:spectral-applications}).
\item \textbf{Experiments.} We test each mechanism with exact finite-library likelihoods on trained image models and a frozen language model (Section~\ref{sec:experiments}).
\end{enumerate}

\paragraph{Summary of the experiments.}
Every sampler we evaluate is a finite library with an exact likelihood, so each comparison is exact for the sampler it specifies. Under uniform retention, training with one or four mixture components gives calibrated likelihoods that agree within seed variation, whereas in a deliberately heterogeneous library output-likelihood training wins all twelve paired comparisons, by $0.32$ to $4.85$ nats per image on average. Retention diagnostics computed from training pixels match the finite-bank expectation of the spectrum, and serial normalization brings the context gain of a frozen language model to the predicted value $1/c_2$ at $16$ rounds (Table~\ref{tab:prediction-tests}).

\paragraph{Organization.}
Section~\ref{sec:setup} sets up the sampler and the local model, and Section~\ref{sec:spectrum} establishes the retention spectrum and the frontier. Section~\ref{sec:uniform} specializes to uniform batching and context correction, and Section~\ref{sec:training} analyzes training with schedule mixtures. Section~\ref{sec:experiments} reports the experiments and Section~\ref{sec:discussion} discusses implications. Proofs, numerical verification and experimental protocols are in the appendices, and Appendix~\ref{app:notation} collects the notation.

\subsection{Related work}
\citet[Theorem~3.3]{chen} compare the divergence of the mixed output with the mean divergence of individual schedules, and \citet{lavenant} and \citet{wainwright} optimize schedules through an information profile and through unmasking growth complexity. \citet{licai} bound the sampling error by the mutual information among tokens, \citet{zhaocai} randomize the number of coordinates revealed per step and obtain rates in total and dual total correlation, and \citet{zhang2026} separate order sensitivity from parallelization bias. \citet{wen} analyze deterministic adaptive reveal trees, and \citet{wu} studies round-complexity separations after mixing. These works bound or optimize the error of a schedule; we describe how a schedule mixture acts on each interaction of the target and what this implies for training and calibration. The ANOVA decomposition and conditional projections are standard tools \citep[Theorem~8.35 and Proposition~8.28]{odonnell}; Exercise~8.17(b) there already allocates a Doob difference by the last revealed coordinate, and Shapley-type variance allocation is studied by \citet{owen}. Our contribution is the identification of the sampler's output derivative and conditional-loss metric, which yields the retention spectrum and its output--calibration frontier. \citet{dependency} characterize conditional updates as information projections and bound stationary full-conditional divergence through conditional KL regrets; our response and loss metric concern the finite-library sampler \eqref{eq:law} and the forward KL of its output distribution.

On the learning side, NADE-k states the mixture-likelihood objective \citep[Eq.~18]{nade}, and DUEL studies exact likelihood under deterministic unmasking \citep{duel}. MDPO and trajectory-aware training address the distribution of training trajectories \citep{mdpo,pumba}. Our finite-mixture analysis quantifies how the standard mixture objective locally interpolates between conditional calibration and inverse retention. Di4C learns mixtures that capture coordinate dependence and distills discrete diffusion models to fewer steps \citep{di4c}, and \citet{fox} learn generation orders through a variational posterior; here the schedule library is fixed and the object of study is the interaction-wise response of a shared scalar kernel. With a fixed reference mask, context-logit extrapolation has the algebraic form of classifier-free guidance, and related reference constructions appear in FoCore and A-CFG \citep{focore,acfg}; the retention spectrum predicts the coefficient for a specified schedule.

\section{Parallel sampling with randomized schedules}\label{sec:setup}
Let $\mathcal X=\prod_{j=1}^d\mathcal X_j$ be finite. A schedule $s$ is an ordered partition $(B_{s,1},\ldots,B_{s,T_s})$, with $T_s\le T$. Let $C_{s,t}=\bigcup_{u<t}B_{s,u}$. A scalar kernel $K_j(x_j\mid x_C,C)$ depends only on previously committed values and their indices, and the same kernel is used whenever the same context occurs, regardless of which schedule is executed. All draws within a batch use the same pre-round history and are independent. Table~\ref{tab:notation} in Appendix~\ref{app:notation} lists the notation.

A fixed finite library has weights $\pi_s>0$ summing to one. Its component and output laws are
\begin{equation}
 q_{s,K}(x)=\prod_t\prod_{j\in B_{s,t}}K_j(x_j\mid x_{C_{s,t}},C_{s,t}),
 \qquad q_{\pi,K}=\sum_s\pi_s q_{s,K}.                 \label{eq:law}
\end{equation}
Only one component is executed at generation time. All logarithms are natural.
\begin{assumption}[Finite system]\label{as:finite}
The alphabets $\mathcal X_j$ and the library $\{(s,\pi_s)\}$ are finite and fixed, $\pi_s>0$, and kernels are shared as described above.
\end{assumption}

For a target $p$, define the schedule-weighted conditional calibration loss, with $\KL$ the Kullback--Leibler (KL) divergence,
\begin{equation}
 \calC_p(K)=\sum_s\pi_s\E_{X\sim p}\sum_t\sum_{j\in B_{s,t}}
 \KL\bigl(p(X_j\mid X_{C_{s,t}})\,\|\,K_j(\cdot\mid X_{C_{s,t}},C_{s,t})\bigr).
 \label{eq:calibration}
\end{equation}
This quantity has a direct training interpretation. Write $K^{\rm or}$ for the target's exact scalar conditionals. A path factorization gives
\begin{equation}
 \sum_s\pi_s\KL(p\|q_{s,K})
 =\sum_s\pi_s\KL(p\|q_{s,K^{\rm or}})+\calC_p(K). \label{eq:pathregret}
\end{equation}
Thus $\calC_p(K)$ is exactly the excess average-path negative log-likelihood.

\paragraph{Local model.}
Fix a full-support product reference $\mu=\bigotimes_j\mu_j$. The target is a twice continuously differentiable positive family
\begin{equation}
 p_\epsilon/\mu=1+\epsilon f+O(\epsilon^2),\qquad \E_\mu f=0. \label{eq:target}
\end{equation}
All spaces and the schedule library are fixed as $\epsilon\to0$, and Gibbs families $p_\epsilon\propto\mu e^{\epsilon f}$ are included. The local analysis uses smooth kernel curves through $K^0_j=\mu_j$ at every visited context, with scores
\begin{equation}
 g_{j,C}(x_j,x_C)=\left.\frac{\partial}{\partial\epsilon}\log K^\epsilon_j(x_j\mid x_C,C)\right|_{\epsilon=0},
 \qquad \E_{\mu_j}g_{j,C}=0,                         \label{eq:score}
\end{equation}
where the last equality holds for each $x_C$. Every such finite score field is attainable by exponentially normalizing $\mu_j e^{\epsilon g_{j,C}}$.
\begin{assumption}[Local model]\label{as:local}
The target family satisfies \eqref{eq:target} around a full-support product law $\mu$, and kernel curves pass through $K^0_j=\mu_j$ at every visited context, with scores \eqref{eq:score}.
\end{assumption}

Let $\alpha_{j,C}$ be the probability that a random schedule draws coordinate $j$ after observing exactly $C$. Ignoring zero-probability contexts, give the score space $\calH_\pi$ the norm
\begin{equation}
 \norm{g}_\pi^2=\sum_{j,C}\alpha_{j,C}\E_\mu g_{j,C}^2. \label{eq:norm}
\end{equation}
The derivative of \eqref{eq:law}, expressed as a density relative to $\mu$, is the linear map
\begin{equation}
 L_\pi g=\sum_{j,C}\alpha_{j,C}g_{j,C}\in L^2_0(\mu). \label{eq:L}
\end{equation}

\section{The retention spectrum and the output--calibration frontier}\label{sec:spectrum}
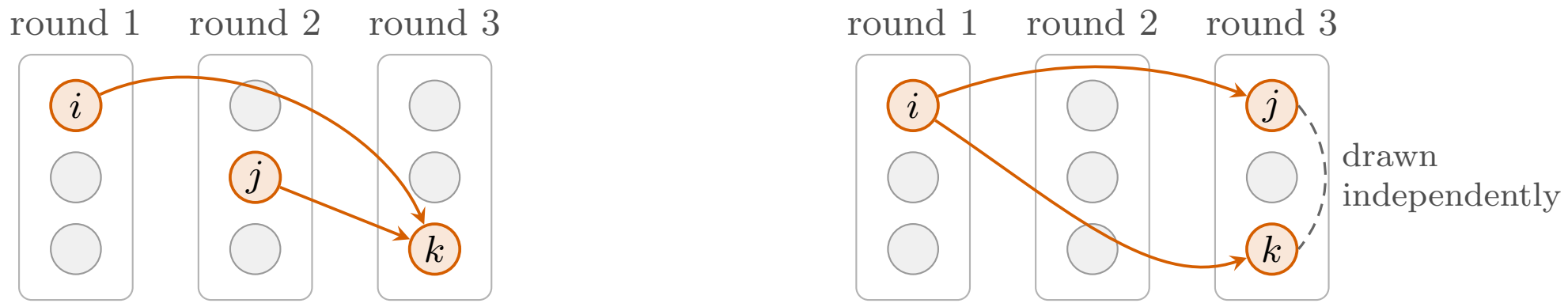
\begin{figure}[t]
\centering
\begin{tikzpicture}[x=1cm,y=1cm,font=\small,
  rnd/.style={draw=black!30,rounded corners=3pt,minimum width=0.95cm,minimum height=2.05cm},
  pt/.style={circle,draw=black!40,fill=black!6,inner sep=0pt,minimum size=4.2mm},
  mem/.style={circle,draw=vermil,fill=vermil!15,line width=0.7pt,inner sep=0pt,minimum size=4.2mm,font=\footnotesize},
  dep/.style={->,>=stealth,vermil,line width=0.7pt}]
\foreach \dx in {0,7.0}{
  \begin{scope}[xshift=\dx cm]
    \foreach \t in {1,2,3}{
      \node[rnd] at ({1.5*(\t-1)},0) {};
      \node[font=\footnotesize,text=black!70] at ({1.5*(\t-1)},1.3) {round \t};
    }
  \end{scope}
}
% (a) unique latest coordinate
\node[mem] (ai) at (0,0.6) {$i$};
\node[pt] at (0,0) {}; \node[pt] at (0,-0.6) {};
\node[pt] at (1.5,0.6) {}; \node[mem] (aj) at (1.5,0) {$j$}; \node[pt] at (1.5,-0.6) {};
\node[pt] at (3,0.6) {}; \node[pt] at (3,0) {}; \node[mem] (ak) at (3,-0.6) {$k$};
\draw[dep] (ai) .. controls (1.2,1.15) and (2.55,0.45) .. (ak);
\draw[dep] (aj) -- (ak);
\node[align=center,font=\footnotesize] at (1.5,-1.6) {(a) unique latest coordinate $k$:\\ $f_A$ is retained};
% (b) tied latest coordinates
\begin{scope}[xshift=7.0cm]
\node[mem] (bi) at (0,0.6) {$i$};
\node[pt] at (0,0) {}; \node[pt] at (0,-0.6) {};
\node[pt] at (1.5,0.6) {}; \node[pt] at (1.5,0) {}; \node[pt] at (1.5,-0.6) {};
\node[mem] (bj) at (3,0.6) {$j$}; \node[pt] at (3,0) {}; \node[mem] (bk) at (3,-0.6) {$k$};
\draw[dep] (bi) .. controls (1.2,1.05) and (2.1,0.95) .. (bj);
\draw[dep] (bi) .. controls (1.2,-0.2) and (2.0,-0.95) .. (bk);
\draw[densely dashed,black!60,line width=0.6pt] (bj.east) to[bend left=40] node[right,font=\scriptsize,text=black!70,align=left] {drawn\\independently} (bk.east);
\node[align=center,font=\footnotesize] at (1.5,-1.6) {(b) $j$ and $k$ tie in the last round:\\ $f_A$ is lost};
\end{scope}
\end{tikzpicture}
\caption{Retention of a three-way interaction $A=\{i,j,k\}$ under one schedule; columns are rounds and arrows show conditioning. (a) The last coordinate $k$ is drawn from a conditional that sees $x_i$ and $x_j$, so its kernel can carry $f_A$. (b) The last two coordinates are drawn independently given the past, so $f_A$ is absent from the output at first order. The retention $c_A$ is the probability of case (a) under the schedule law $\pi$.}
\label{fig:retention}
\end{figure}

Write $f=\sum_{\varnothing\ne A\subseteq[d]}f_A$ for the orthogonal product-space interaction decomposition. Each $f_A$ depends on $x_A$ and has mean zero in every coordinate of $A$ while the others are fixed. Let $P_A$ be its orthogonal projection. If $t_s(j)$ is the round of coordinate $j$, define
\begin{equation}
 c_A=\Prb_{s\sim\pi}\!\left\{\max_{j\in A}t_s(j)\text{ has a unique maximizer}\right\},
 \qquad R_\pi=\sum_{A\ne\varnothing}c_AP_A.          \label{eq:retention}
\end{equation}
For a pair $\{i,j\}$, $c_{ij}$ is the probability that $i$ and $j$ are revealed in different rounds. Singletons always have retention one. We call an interaction \emph{active} if $f_A\ne0$ and \emph{covered} if $c_A>0$. Figure~\ref{fig:retention} illustrates the definition.

\begin{example}\label{ex:retention}
Let $d=3$ and let $\pi$ give equal weight to the schedules $(\{0,2\},\{1\})$ and $(\{0\},\{1,2\})$, the library used again in Section~\ref{sec:three-spin}. Coordinate $1$ is the unique latest member of $\{0,1\}$ in both schedules, so $c_{01}=1$. The pair $\{0,2\}$ ties in the first schedule, and the pair $\{1,2\}$ and the triple $\{0,1,2\}$ tie in the second, so $c_{02}=c_{12}=c_{012}=1/2$. Exact conditionals thus keep the interaction of $x_0$ and $x_1$ intact and halve the other three.
\end{example}

\begin{theorem}[Retention spectrum and sharp output--calibration frontier]\label{thm:main}
Under Assumptions~\ref{as:finite} and~\ref{as:local}, the following hold.

\textup{(i)} \emph{Spectrum.} The adjoint of $L_\pi$ and the response Gram operator satisfy
\begin{equation}
 (L_\pi^*f)_{j,C}=\E_\mu[f\mid X_{C\cup\{j\}}]-\E_\mu[f\mid X_C],
 \qquad \boxed{L_\pi L_\pi^*=R_\pi}.                \label{eq:spectrum}
\end{equation}
\textup{(ii)} \emph{Local risks.} The exact conditional predictor has score $g^{\rm or}=L_\pi^*f$, and for any differentiable kernel curve with score $g$ the output and calibration coefficients are
\begin{align}
 \lim_{\epsilon\to0}\frac{2\KL(p_\epsilon\|q_{\pi,K^\epsilon})}{\epsilon^2}
   &=\norm{f-L_\pi g}_{L^2(\mu)}^2, \label{eq:outputrisk}\\
 \lim_{\epsilon\to0}\frac{2\calC_{p_\epsilon}(K^\epsilon)}{\epsilon^2}
   &=\norm{g-g^{\rm or}}_\pi^2.                       \label{eq:calrisk}
\end{align}
\textup{(iii)} \emph{Frontier.} For every $B\ge0$, the smallest output coefficient attainable with calibration coefficient at most $B$ is
\begin{equation}
 \boxed{\min_{\substack{u_A\in\operatorname{range}P_A\ (c_A>0)\\
                    \sum_{c_A>0}\norm{u_A}^2/c_A\le B}}
 \left\{\sum_{c_A>0}\norm{(1-c_A)f_A-u_A}^2
          +\sum_{c_A=0}\norm{f_A}^2\right\}.}        \label{eq:frontier}
\end{equation}
Here all unmarked norms are in $L^2(\mu)$. In particular, exact conditional sampling has output coefficient
\begin{equation}
 \sum_A(1-c_A)^2\norm{f_A}^2.                        \label{eq:oracle}
\end{equation}
Over smooth local kernel curves, the unrestricted smallest coefficient is $\sum_{c_A=0}\norm{f_A}^2$. The minimum calibration coefficient attaining it is
\begin{equation}
 B_*=\sum_{c_A>0}\frac{(1-c_A)^2}{c_A}\norm{f_A}^2.   \label{eq:bstar}
\end{equation}
When every active interaction has positive retention, this optimum is achieved by shared kernels with $\KL(p_\epsilon\|q_{\pi,K^\epsilon})=O(\epsilon^4)$.
\end{theorem}

The theorem gives a necessary and sufficient condition for recovering the target to first order with shared kernels, together with the exact calibration cost of doing so. The spectrum is independent of the target; interaction magnitudes enter only when the frontier is evaluated. Small retention therefore matters even for covered interactions: it makes a recovered interaction more expensive in conditional-prediction loss, at the rate $1/c_A$.

\paragraph{Proof sketch.}
The adjoint is the conditional projection that adds one coordinate to a visible history. On an interaction, that projection contributes once exactly when the schedule has a unique latest coordinate, and averaging these contributions gives the spectrum. Quadratic KL expansion identifies the output and conditional-loss metrics; minimum-norm lifting through the response operator then gives the frontier and its attaining kernels. Appendix~\ref{app:main-theorem-proof} gives the complete proof and the extension to positive kernel sequences with bounded calibration cost.

\subsection{Regularization traverses the entire frontier}
\begin{figure}[t]
\centering\includegraphics{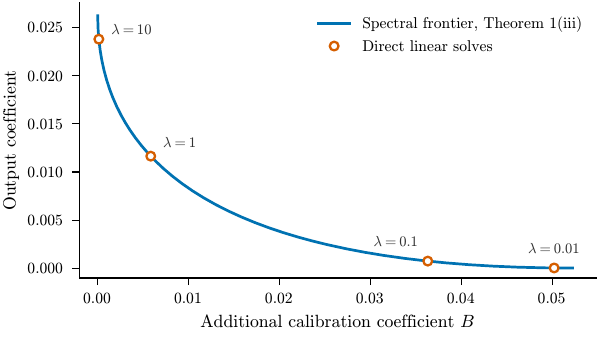}
\caption{The output--calibration frontier of Theorem~\ref{thm:main}(iii) for the biased four-coordinate binary example of Appendix~\ref{app:numerical-verification}. Circles are the minimizers of the regularized objective at $\lambda\in\{0.01,0.1,1,10\}$, computed by direct linear solves; the curve is the spectral frontier \eqref{eq:frontier}. Both axes are doubled local KL coefficients.}
\label{fig:calibration-frontier}
\end{figure}

\begin{corollary}[Closed-form regularized correction]\label{cor:regularized}
For $\lambda>0$, minimizing the leading coefficient of
$\KL(p_\epsilon\|q_{\pi,K})+\lambda\calC_{p_\epsilon}(K)$ gives
\begin{align}
 g_\lambda
 &=L_\pi^*f+L_\pi^*(R_\pi+\lambda I)^{-1}(I-R_\pi)f, \label{eq:ridge}\\
 (g_\lambda)_{j,C}
 &=\sum_{\substack{A\ni j\\A\setminus\{j\}\subseteq C}}
       \frac{1+\lambda}{c_A+\lambda}f_A.            \label{eq:ridgefield}
\end{align}
The minimum doubled coefficient is
\begin{equation}
 \sum_A\frac{\lambda(1-c_A)^2}{c_A+\lambda}\norm{f_A}^2, \label{eq:ridgevalue}
\end{equation}
with calibration coefficient
$B_\lambda=\sum_A c_A(1-c_A)^2\norm{f_A}^2/(c_A+\lambda)^2$.
\end{corollary}
\begin{proof}
Minimize $\norm{r-L_\pi h}^2+\lambda\norm{h}_\pi^2$.
The normal equations give $h=L_\pi^*(R_\pi+\lambda I)^{-1}r$, and substitution in each interaction space gives the stated values. As $\lambda$ decreases from infinity to zero, $B_\lambda$ increases continuously from zero to $B_*$; degenerate zero-cost cases follow by continuity.
\end{proof}
Figure~\ref{fig:calibration-frontier} traces this path for a four-coordinate binary example: the regularized solutions, computed by direct linear solves, lie on the spectral frontier (Appendix~\ref{app:numerical-verification}).

The corresponding observable losses are
\begin{align*}
 \calL_{\rm out}(K)&=-\E_p\log q_{\pi,K}(X),\\
 \calL_{\rm path}(K)&=-\E_p\sum_s\pi_s\log q_{s,K}(X).
\end{align*}
By \eqref{eq:pathregret}, minimizing
\begin{equation}
 \calL_{\rm out}(K)+\lambda\calL_{\rm path}(K)       \label{eq:train}
\end{equation}
has the same $K$-dependent terms as output KL plus $\lambda\calC_p(K)$. Both losses are estimated from observed examples, without conditional-probability labels. The theorem characterizes their optimum in score space; in a constrained model it provides a correction direction or a regularization target.

\subsection{An explicit finite-perturbation guarantee}
\begin{proposition}[Finite radius for the unregularized correction]\label{prop:finite}
Suppose that every active interaction is covered, and let $p_\epsilon=\mu e^{\epsilon f}/\E_\mu e^{\epsilon f}$ with $\E_\mu f=0$. Set $F=\norm{f}_\infty$ and $G=\max_{j,C}\norm{g^*_{j,C}}_\infty$. If
$|\epsilon|\max(F,dG)\le1/4$, then exponentially normalized kernels with the scores $g^*$ of \eqref{eq:gstar} satisfy
\begin{equation}
 \KL(p_\epsilon\|q_{\pi,K})
 \le 8\epsilon^4\bigl(F^2+d^2G^2\bigr)^2.           \label{eq:finite}
\end{equation}
\end{proposition}
The proof is in Appendix~\ref{app:finite}. The proposition gives an explicit radius and remainder for a specified finite target and library.

\paragraph{Further consequences.}
The same spectrum gives the exact leading error of the best deterministic adaptive sampler on Ising models, a convex objective for schedule weights, and an interaction correction estimated from samples, with a finite-sample guarantee. Appendix~\ref{app:spectral-applications} develops these consequences with complete proofs.

\section{Retention under uniform batching and context correction}\label{sec:uniform}
\subsection{Retention under uniform fixed-quota batching}
\label{sec:quota-retention}
Sample a uniform permutation of the $d$ coordinates and split it into batches of fixed sizes $m_1,\ldots,m_T$, with $\sum_t m_t=d$. Write $a_t=\sum_{u<t}m_u$. Exchangeability makes the retention coefficient depend only on the interaction order $k=|A|$:
\begin{equation}
 c_k=\frac{\sum_{t=1}^T m_t\binom{a_t}{k-1}}{\binom{d}{k}},
 \label{eq:quota-ck}
\end{equation}
where a binomial coefficient is zero when its lower argument exceeds its upper argument. Indeed, the positions occupied by $A$ form a uniform $k$-subset of the $d$ permutation positions; a unique latest position in batch $t$ can be chosen in $m_t$ ways, and the other $k-1$ positions must lie in earlier batches. In particular, for $d\ge2$,
\begin{equation}
 c_2=\frac{d^2-\sum_t m_t^2}{d(d-1)}.
 \label{eq:quota-c2}
\end{equation}
For equal batches $m=d/T$, this becomes $c_2=(d-m)/(d-1)$; for single-coordinate batches it gives $c_2=1$. Fixed quotas differ from assigning each coordinate an independent round label: with label probabilities $p_1,\ldots,p_T$, one gets $c_k=k\sum_t p_t(\sum_{u<t}p_u)^{k-1}$ and $c_2=1-\sum_t p_t^2$. Equation~\eqref{eq:quota-c2} is the coefficient for a random permutation cut into prescribed batches, including unequal batches when $T$ does not divide $d$.

\subsection{A local context-gain corollary}
\label{sec:local-context-gain}
Consider binary coordinates and the local model of Theorem~\ref{thm:main}, with
\begin{equation}
 f=\sum_j f_{\{j\}}+\sum_{i<j}f_{\{i,j\}}.
 \label{eq:pairwise-local-score}
\end{equation}
Let $\ell^\epsilon_{j,C}(x_C)$ denote the logit of the target conditional $p_\epsilon(X_j=1\mid x_C)$, including its full finite-$\epsilon$ dependence. For a scalar $\gamma$, define a shared kernel by
\begin{equation}
 \widetilde K^\epsilon_{j,\gamma}(1\mid x_C,C)
 =\operatorname{sigmoid}\!\left(
 \ell^\epsilon_{j,\varnothing}
 +\gamma\bigl[\ell^\epsilon_{j,C}(x_C)
                   -\ell^\epsilon_{j,\varnothing}\bigr]\right).
 \label{eq:context-gain}
\end{equation}
At first order, the all-masked logit carries the singleton contribution, while the logit difference carries the contextual interactions.

\begin{corollary}[Pairwise context gain]\label{cor:context-gain}
Under~\eqref{eq:pairwise-local-score}, the output of
\eqref{eq:context-gain} satisfies
\begin{align}
 \frac{q_{\pi,\widetilde K^\epsilon_\gamma}}{\mu}
 &=1+\epsilon\left(\sum_j f_{\{j\}}
              +\gamma\sum_{i<j}c_{ij}f_{\{i,j\}}\right)
              +O(\epsilon^2),\label{eq:gain-response}\\
 \lim_{\epsilon\to0}\frac{2\KL(p_\epsilon\|
            q_{\pi,\widetilde K^\epsilon_\gamma})}{\epsilon^2}
 &=\sum_{i<j}(1-\gamma c_{ij})^2\norm{f_{\{i,j\}}}^2.
 \label{eq:gain-local-risk}
\end{align}
If all active pairs have the same $c_{ij}=c>0$, then $\gamma=1/c$ gives
$\KL(p_\epsilon\|q_{\pi,\widetilde K^\epsilon_\gamma})=O(\epsilon^4)$
and attains the minimum additional calibration coefficient
\begin{equation}
 B_* =\frac{(1-c)^2}{c}\sum_{i<j}\norm{f_{\{i,j\}}}^2.
 \label{eq:gain-mincost}
\end{equation}
The regularized frontier on these interactions is attained by
$\gamma_\lambda=(1+\lambda)/(c+\lambda)$, for $\lambda>0$.
\end{corollary}
\begin{proof}
The score of the all-masked kernel is the singleton oracle score, and subtracting it from a contextual logit removes that singleton term to first order. Consequently the kernel score in~\eqref{eq:context-gain} is
$L_\pi^*(\sum_j f_{\{j\}}+\gamma\sum_{i<j}f_{\{i,j\}})$.
Apply $L_\pi L_\pi^*=\sum_Ac_AP_A$, with $c_{\{j\}}=1$, to obtain
\eqref{eq:gain-response}; orthogonality gives~\eqref{eq:gain-local-risk}.
At $\gamma=1/c$ the first-order output residual vanishes, and the finite-state second-order density remainder gives fourth-order KL. The additional score has squared norm
$(\gamma-1)^2\sum_{i<j}c_{ij}\norm{f_{\{i,j\}}}^2$,
which yields~\eqref{eq:gain-mincost} and the regularized coefficient from Theorem~\ref{thm:main}.
\end{proof}

For fixed-quota uniform batching, the predicted gain is therefore $\gamma=1/c_2$, with $c_2$ from~\eqref{eq:quota-c2}. For heterogeneous pair retention, the best scalar gain in the local output criterion is
\begin{equation}
 \gamma^*_{\rm loc}=
 \frac{\sum_{i<j}c_{ij}\norm{f_{\{i,j\}}}^2}
      {\sum_{i<j}c_{ij}^2\norm{f_{\{i,j\}}}^2},
 \label{eq:weighted-scalar-gain}
\end{equation}
when the denominator is positive. Interaction-specific correction can recover covered pairs that a scalar gain cannot simultaneously restore, and higher-order interactions, whose retentions generally differ, are not all corrected by $1/c_2$. Section~\ref{sec:experiments} applies \eqref{eq:context-gain} to learned logits. When every all-masked logit is zero, context gain coincides with global temperature; Appendix~\ref{app:remarks} discusses the general relation, and Appendix~\ref{app:gain-displacement} shows that the optimal gain moves linearly with the coupling at finite strength.

\section{Training with schedule mixtures}\label{sec:training}
\subsection{Finite schedule mixtures}
\label{sec:finiteM}
Draw $S_1,\ldots,S_M$ independently from the fixed schedule law $\pi$, independently of the datum, and let
\begin{equation}
 \widehat q_{M,K}=\frac1M\sum_{m=1}^M q_{S_m,K},\qquad
 \calL_M(K)=\E_{X\sim p}\E_{S_{1:M}}[-\log\widehat q_{M,K}(X)].
 \label{eq:finiteM-loss}
\end{equation}
For positive finite-state kernels, Jensen's inequality and a leave-one-out argument give
\begin{equation}
 \calL_{\rm out}(K)\le\calL_{M+1}(K)\le\calL_M(K),
 \qquad \lim_{M\to\infty}\calL_M(K)=\calL_{\rm out}(K).
 \label{eq:finiteM-monotonic}
\end{equation}
These are the importance-weighted likelihood bounds of \citet[Theorem~1]{iwae}, with the schedule as the latent variable and its prior as the proposal. At $M=1$, $\calL_1=\calL_{\rm path}$ for the specified reveal law and its visitation weights.

\begin{theorem}[Finite-mixture training on the local frontier]
\label{cor:finiteM}
Under Assumptions~\ref{as:finite} and~\ref{as:local}, with the integer $M$ fixed as $\epsilon\to0$, put
\begin{equation}
 J_M(g)=\lim_{\epsilon\to0}
 \frac{2\E_{S_{1:M}}\KL(p_\epsilon\|\widehat q_{M,K^\epsilon})}{\epsilon^2}.
 \label{eq:finiteM-coefficient}
\end{equation}
Then
\begin{equation}
 J_M(g)=\norm{f-L_\pi g}^2
       +\frac1M\bigl(\norm{g}_\pi^2-\norm{L_\pi g}^2\bigr).
 \label{eq:finiteM-variance}
\end{equation}
For $M>1$, minimizing the kernel-dependent local coefficient of
$\calL_M+\lambda\calL_{\rm path}$, $\lambda\ge0$, selects the same frontier
point as output KL plus $\lambda_{\rm eff}\calC_p$, where
\begin{equation}
 \lambda_{\rm eff}=\frac{M\lambda+1}{M-1}.
 \label{eq:finiteM-effective-lambda}
\end{equation}
Its optimal shared-kernel score, with $C$ the set of coordinates revealed before the current round, is
\begin{equation}
 (g_{M,\lambda})_{j,C}=
 \sum_{\substack{A\ni j,\ A\setminus\{j\}\subseteq C\\c_A>0}}
 \frac{1+\lambda}{(1-1/M)c_A+1/M+\lambda}\,f_A.
 \label{eq:finiteM-optimum}
\end{equation}
For $M=1$ the optimum is the conditional-oracle score. An interaction with $c_A=0$
remains unrecoverable to first order.
\end{theorem}
\emph{Proof idea.} Along one schedule the visited score fields are orthogonal, so an $M$-sample mixture has output score $L_\pi g$ plus a fluctuation whose expected squared norm is $(\norm{g}_\pi^2-\norm{L_\pi g}^2)/M$. The identity $L_\pi L_\pi^*=R_\pi$ turns this term into a multiple of the calibration loss, and Corollary~\ref{cor:regularized} then gives the optimum; Appendix~\ref{app:finiteM} has the details. For $\lambda=0$, finite $M$ gives partial correction with coefficient $M/[1+(M-1)c_A]$, and full inverse retention is its $M\to\infty$ limit. At the population score-space optimum, when all pairs share one retention $c>0$, the additional context gain that restores the remaining first-order attenuation is
\begin{equation}
 \gamma_{\rm residual}(M)=1+\frac{1-c}{Mc}.
 \label{eq:finiteM-residual-gain}
\end{equation}
At $\lambda=0$, a covered interaction retains the fraction $Mc_A/(1-c_A+Mc_A)$ of its target amplitude, so interactions with small retention remain strongly attenuated at small $M$.

\subsection{When output training beats any scalar}
\label{sec:calibration-geometry}
Theorem~\ref{cor:finiteM} also answers a practical question: when can the
change that likelihood training makes to the interaction field be
reproduced by calibrating the model after training? We use its population
objective with $\lambda=0$, independent schedule draws for each datum and
fixed $M$ as $\epsilon\to0$, for binary coordinates and the
singleton--pair score~\eqref{eq:pairwise-local-score}. All risks $D$ below
are coefficients of $2\KL/\epsilon^2$ in this fixed finite system.

Write $E_A=\norm{f_A}^2$ and restrict unqualified sums to pairs with
$c_A>0$ and $E_A>0$. Let $E_0$ be the energy of zero-retention pairs
and $S_r=\sum_A c_A^rE_A$. A context gain preserves the singleton field
and multiplies every pair score by the same scalar. If $S_2>0$, its
optimal output coefficient when applied to the conditional oracle is
\begin{equation}
 D_{\rm scalar}=E_0+S_0-\frac{S_1^2}{S_2},\qquad
 \gamma_{\rm scalar}=\frac{S_1}{S_2}.                 \label{eq:best-scalar}
\end{equation}
The excess over $E_0$ vanishes exactly when all active pairs with positive
retention share a single retention level.

Heterogeneous retention arises in common decoding strategies.
Semi-autoregressive block decoding, used for instance in LLaDA
\citep{llada}, generates blocks of coordinates from left to right. With a
random order inside each block, pairs in different blocks have retention
one, whereas a pair inside a block of $b$ coordinates revealed in rounds of
sizes $m_1,\ldots,m_n$ has retention $(b^2-\sum_tm_t^2)/(b(b-1))$ by
\eqref{eq:quota-c2}, which is below one as soon as some round reveals two
coordinates of the block.

For the population finite-$M$ optimum in the full score space, define
$a_M(c)=Mc/[1+(M-1)c]$. Its output coefficient before calibration is
\begin{equation}
 D_{M,\rm raw}=E_0+\sum_A
       \frac{(1-c_A)^2}{[1+(M-1)c_A]^2}E_A.          \label{eq:raw-finiteM}
\end{equation}
Because finite $M$ corrects each interaction only partially, this raw coefficient
can exceed the best scalar coefficient even under heterogeneous retention
(Appendix~\ref{app:remarks} gives a two-pair example). The informative
comparison therefore applies the same scalar after training.

\begin{corollary}[Finite-mixture training followed by the same scalar]
\label{cor:postscalar}
Under the preceding assumptions, with $S_2>0$, optimizing one context
gain after the full score-space population optimum gives
\begin{equation}
 D_{M,\rm scaled}=E_0+S_0-
 \frac{[\sum_A E_A a_M(c_A)]^2}{\sum_A E_A a_M(c_A)^2}
 \le D_{\rm scalar}.                              \label{eq:postscalar}
\end{equation}
For $M>1$, the inequality is strict if and only if at least two distinct positive
retentions carry nonzero interaction energy. If all such retentions
equal $c$, the trained field already belongs to the scalar family;
its coefficient is $M/[1+(M-1)c]$, reaching $1/c$ only as $M\to\infty$
unless $c=1$.
\end{corollary}
\begin{proof}
Scalar least squares gives both displayed risks. Normalize the positive
interaction energies to a probability law and let $C$ denote retention.
For $r_t=C/(1+tC)$, set $H(t)=(\E r_t)^2/\E r_t^2$. Direct differentiation
gives
\[
 \frac{d}{dt}\log H(t)
 =2\left(\frac{\E r_t^3}{\E r_t^2}
           -\frac{\E r_t^2}{\E r_t}\right)\ge0,
\]
by Cauchy--Schwarz, strictly when $C$ is nonconstant. Taking $t=M-1$
proves the claim; the common factor $M$ does not affect $H$.
\end{proof}

\paragraph{The heterogeneity gap.}
In the same local coefficient units, define
\begin{equation}
 H=S_0-\frac{S_1^2}{S_2}.
 \label{eq:heterogeneity-budget}
\end{equation}
Corollary~\ref{cor:postscalar} gives the immediate consequence
\begin{equation}
 0\le D_{\rm scalar}-D_{M,\rm scaled}\le H,\qquad
 \lim_{M\to\infty}(D_{\rm scalar}-D_{M,\rm scaled})=H.
 \label{eq:heterogeneity-limit}
\end{equation}
Indeed, $a_M(c)\to1$ for every active positive retention, so the
scaled and raw optima tend to the uncovered energy $E_0$, and every
scaled risk is at least $E_0$. Thus $H$ is exactly the improvement over the best scalar at the
output-likelihood population optimum, and it bounds the improvement of
finite-mixture training followed by the best scalar for every $M$ (the
limit in $M$ is taken after $\epsilon\to0$). In nats, an improvement of
$H$ corresponds to $\epsilon^2H/2+o(\epsilon^2)$.

\paragraph{Richer calibration families.}
Our experiments also use the two-parameter calibration
$\alpha\ell_{\varnothing}+\gamma(\ell_C-\ell_{\varnothing})$, which
multiplies the all-masked logit by a marginal coefficient $\alpha$ and the
contextual part by the gain $\gamma$. On curves through the same product
reference, $\alpha$ changes only singleton score directions, and the pair
field still has the single common factor $\gamma$, so this family reaches
no additional pair directions. More generally,
if $b_r$ are the available calibration score directions, recovery of the
oracle's missing output field requires that field to lie in
$\operatorname{span}\{L_\pi b_r\}$. In a pair predictor with one weight per
interaction, dividing each weight of the exact first-order oracle by its
retention attains the local optimum on covered interactions without
changing the training objective.

\subsection{Finite banks and serial normalization}
\label{sec:finite-banks}
\paragraph{Finite random banks.}
A \emph{bank} is a library of $K$ schedules drawn independently from the
uniform fixed-quota law. Let a bank be independent of fixed pair energies
$E_{ij}\ge0$ with positive sum.
Put $w_{ij}=E_{ij}/\sum E_{ij}$, $c=c_2>0$ from
\eqref{eq:quota-c2}, and let $\widehat c_{ij}$ be the bank's empirical
pair retention. Each pair averages $K$ independent Bernoulli$(c)$
indicators. Therefore its normalized oracle residual satisfies
\begin{equation}
 \E R_K
 :=\E\sum_{i<j}w_{ij}(1-\widehat c_{ij})^2
 =(1-c)^2+\frac{c(1-c)}K .
 \label{eq:finite-bank-residual}
\end{equation}
No independence between distinct pairs is needed. A random bank thus pays
a heterogeneity premium of exactly $c(1-c)/K$ over the full uniform law,
and banks of a few hundred schedules are already close to homogeneous;
Appendix~\ref{app:remarks} bounds the expected heterogeneity gap of such
banks in the same way.

\paragraph{Normalizing by a serial reference.}
A learned predictor can carry a calibration error that is common to
parallel and serial revealing, and comparison with serial revealing
removes it. Suppose that, at all histories visited by both samplers, a
learned predictor has no other first-order score components, has the
oracle singleton score and has a pair score equal to $\beta>0$ times the
oracle pair score, with the same $\beta$ for the $T$-round library and
serial revealing. For $S_2>0$, scalar least squares then has unique
minimizers
\begin{equation}
 \gamma_T^*=\frac{S_1}{\beta S_2},\qquad
 \gamma_0^*=\frac1\beta,\qquad
 \frac{\gamma_T^*}{\gamma_0^*}=\frac{S_1}{S_2}.
 \label{eq:serial-normalization}
\end{equation}
Indeed, the pair risks are $\sum_A(1-\gamma\beta c_A)^2E_A$
and $\sum_A(1-\gamma\beta)^2E_A$, respectively, with zero-retention
terms included in these sums. When every active pair has retention
$c>0$, the ratio is $1/c$, whatever the common factor $\beta$. For
heterogeneous retention it is the energy-weighted moment ratio
in~\eqref{eq:serial-normalization}, rather than the reciprocal of mean
retention. Appendix~\ref{app:remarks} extends the identity to
interaction-dependent factors.

\subsection{A certified advantage at finite coupling}
\label{sec:three-spin}
Within a fixed library, the advantage of output training persists away
from the asymptotic coefficients. For three spins let $y=x_0x_1$,
$z=x_0x_2$ and $p_t(x)=\frac18(1+ty)(1+tz)$ with $t=1/10$. Two
equal-weight schedules, $(\{0,2\},\{1\})$ and $(\{0\},\{1,2\})$, have pair
retentions $c_{01}=1,c_{02}=1/2$. A shared logistic predictor with two
edges, with edge means $u,v\in(-1,1)$, gives
\[
 q_A=\tfrac18(1+uy),\quad
 q_B=\tfrac18(1+uy)(1+vz),\quad
 q_\pi=\tfrac18(1+uy)(1+vz/2).
\]
At every visited history the exact conditional oracle is $u=v=t$.
Its empty-context logits are zero, so the two-parameter calibration
reduces exactly to one scalar, with $u=v=w$ ranging over $(-1,1)$.
Define the binary KL between means $t$ and $a$ by
\[
 C_t(a)=\frac12\sum_{\sigma\in\{-1,1\}}(1+\sigma t)
          \log\frac{1+\sigma t}{1+\sigma a}.
\]
The best scalar minimizes $C_t(w)+C_t(w/2)$. Finite-$M$
population training minimizes $C_t(u)+\E C_t(Zv)$, where
$Z\sim\mathrm{Binomial}(M,1/2)/M$. These objectives are strictly convex.
The training optimum has $u=t$ and the unique root $v\in(-1,1)$ of
\[
 \E\!\left[\frac{Z(Zv-t)}{1-Z^2v^2}\right]=0,
\]
and the output KL of the trained sampler is $C_t(v/2)$.
Rational root isolation and a log-series remainder bound certify best
scalar KL $0.0010069871\ldots$, versus $0.0005639657\ldots$ for $M=2$
and $0.0002054298\ldots$ for $M=4$. The certified improvements exceed
$4.43\cdot10^{-4}$ and $8.015\cdot10^{-4}$ nats, respectively: training
with four schedule draws reduces the best scalar's KL by a factor of
almost five. As $M\to\infty$, $Z\to1/2$ and the training optimum tends to
$v=2t$, the edge-specific correction, which attains zero KL.

\section{Experiments}\label{sec:experiments}
We now test the mechanisms of Sections~\ref{sec:spectrum}--\ref{sec:training}
on trained predictors. Three features of the design make the tests sharp.
First, every sampler we evaluate is a fixed finite library, whose output
likelihood is an exact sum over its components, so a difference between two
objectives is a difference between two exactly specified samplers, free of
Monte Carlo integration error. Second, each comparison holds the library
fixed and changes one thing: the training objective, the calibration or
the library. Third, the pairwise logistic field implements exactly the
first-order field of the theory, while the MLP shows how the mechanisms
carry over to a generic predictor. Image distributions are far from product laws, so the
experiments also probe how well the mechanisms hold beyond the local
regime of the theory; full-state calculations confirm the response
spectrum, the frontier and the results of Appendix~\ref{app:spectral-applications}
to numerical precision (Appendix~\ref{app:numerical-verification}).
Table~\ref{tab:prediction-tests} summarizes the tests.

\begin{table}[!b]
\centering\small
\caption{Predictions of the theory and their tests. Predictions concern the
local population optimum. Differences are in nats/image, with training-seed
standard deviations.}
\label{tab:prediction-tests}
\begin{tabular}{@{}>{\raggedright\arraybackslash}p{0.31\textwidth}>{\raggedright\arraybackslash}p{0.27\textwidth}>{\raggedright\arraybackslash}p{0.36\textwidth}@{}}
\toprule
Prediction & Test & Result\\
\midrule
Response spectrum $L_\pi L_\pi^*=R_\pi$; fourth-order KL after inverse
retention (Theorem~\ref{thm:main})
& Full-state computation (Appendix~\ref{app:numerical-verification})
& $\max|L_\pi L_\pi^*-R_\pi|<10^{-15}$; log-KL slopes $2.0$ for exact
conditionals and $4.0$ after inverse retention.\\[3pt]
Homogeneous retention: training adds nothing beyond one scalar
(Corollary~\ref{cor:postscalar})
& Calibrated $M=4$ minus $M=1$, uniform library
& Digits: $-0.0006\pm0.0447$.
MNIST (full training budget): $-0.068\pm0.102$.\\[3pt]
Heterogeneous retention: training gains beyond any scalar, up to $H$
\eqref{eq:heterogeneity-limit}
& Output versus path training in a two-block library
& Output wins all 12 paired comparisons; mean gains from $0.32$ to
$4.85$.\\[3pt]
The output-optimal pair field needs unit residual gain
& Gain selected with marginal coefficient fixed at one
& Pair field, path $\to$ output training: $1.225\to1.008$ (digits),
$1.100\to1.025$ (MNIST).\\[3pt]
Finite-bank residual $(1-c)^2+c(1-c)/K$
\eqref{eq:finite-bank-residual}
& Training-pixel correlation proxies on the evaluated banks
& MNIST $K=18/128/512$: observed 0.0726/0.0635/0.0624,
predicted 0.0724/0.0635/0.0624.\\[3pt]
Serial normalization $\gamma_T/\gamma_0=1/c$
\eqref{eq:serial-normalization}
& Frozen 169.6M-parameter MDLM, two banks and two serial orders
& $1.070/1.056$ at $T=8/16$, against $1/c_2=1.134/1.058$.\\
\bottomrule
\end{tabular}
\end{table}

\subsection{Setup}
We train a symmetric masked logistic field and a two-hidden-layer,
width-$256$ GELU multilayer perceptron (MLP) on binarized $8\times8$ digits
($d=64$) and MNIST ($d=784$), and report negative log-likelihood (NLL) in
nats. The logistic field has a learned bias and symmetric pair
weights acting on the observed pixels; the MLP receives the revealed
pixels as $\pm1$ values, with zeros elsewhere, and visibility indicators. Both start from random-mask
pretraining. In the uniform-library study the MLP is fine-tuned with one
or four schedule draws per example; in the heterogeneous-library study
both models are fine-tuned with the exact mean-path or output likelihood
of the library. Validation selects checkpoints and logit corrections of
three kinds: the context gain~\eqref{eq:context-gain}, a global
temperature, and the two-parameter correction
$\alpha\ell_{\varnothing}+\gamma(\ell_C-\ell_{\varnothing})$ of
Section~\ref{sec:calibration-geometry}.
Appendices~\ref{app:observed-protocol} and~\ref{app:extension-protocol}
give the data splits, optimization, selection rules and cost accounting.

\subsection{Uniform retention: one scalar suffices}
\label{sec:learned}
The full uniform permutation law has homogeneous pair retention $c_2$,
so Corollary~\ref{cor:postscalar} predicts that finite-mixture training
changes the output law only along the direction that a single context
gain also reaches. At $T=4$ we fine-tune the MLP from the same base
checkpoint with $M=1$ or $M=4$ resampled schedule components per example,
using five seeds, and evaluate both on two fixed libraries of $K=512$
uniform schedules.

\paragraph{Training moves along the output--calibration frontier.}
Without calibration, $M=4$ lowers exact output NLL relative to $M=1$ by
$0.215\pm0.037$ nats/image on digits and $0.495\pm0.251$ on MNIST, while
raising mean path NLL by $0.673\pm0.033$ and $1.675\pm0.200$; each
direction holds in all five seeds. By~\eqref{eq:pathregret}, the path
difference is exactly the difference in conditional calibration loss, so
the trained models trade conditional calibration for output accuracy, in
the direction predicted by Theorem~\ref{cor:finiteM}
(Figure~\ref{fig:path-output-plane} in Appendix~\ref{app:uniform-controls}).

\paragraph{A scalar recovers the difference.}
After each model is recalibrated on the evaluated banks with the
two-parameter correction, the paired $M=4$ minus $M=1$ difference is
$-0.0006\pm0.0447$ nats/image on digits. On MNIST, models trained to
the full update budget and scored on a separate 512-image test range give
$-0.068\pm0.102$ (Appendix~\ref{app:extension-protocol};
Appendix~\ref{app:uniform-controls} reports the earlier, shorter runs). Even the predicted gain $1/c_2$
applied to the $M=1$ model outperforms the uncalibrated $M=4$ model in
every seed of both data sets, and the randomized library outperforms two
separately trained fixed schedules with matched training budgets in every
seed.

\paragraph{Serial normalization.}
For the same MNIST $M=1$ models, the mean serially normalized ratios
(selected context gain divided by the gain selected under serial revealing)
are $1.441$ and $1.396$ for two fixed serial orders, $8.2\%$ and $4.8\%$
above the predicted value $1/c_2=1.332$.

\subsection{Heterogeneous retention: training beyond a scalar}
\label{sec:heterogeneous-learning}
We next build a library in which positive retentions differ, the
\emph{two-block library}. Split the pixels into top and bottom halves and give the pixels of each half
distinct $b$-bit binary codes. Each bit position defines two schedules:
one reveals the pixels whose bit is $0$ before those whose bit is $1$, and
the other reverses this order. The top half occupies rounds 1 and 2 and the
bottom half rounds 3 and 4, so the library has $K=2b$ components.
Cross-half pairs have retention 1, and a within-half pair has retention
equal to the Hamming distance of its two codes divided by $b$. Digits uses $(d,b,K)=(64,5,10)$ and MNIST
$(784,9,18)$. All pair retentions are positive, so every pair is
recoverable, but to different degrees.

Both models start from the same checkpoint for each seed (11, 22 and 33)
and optimize the exact mean-path or output NLL of this library,
evaluating all components at every update; output NLL is the
$M\to\infty$ limit of Section~\ref{sec:finiteM} at fixed library size $K$.
Validation selects checkpoints and logit corrections, and the MNIST test
set of 512 previously unused images is fixed before training.

\begin{table}[t]
\centering\small
\caption{Heterogeneous-library output NLL after validation-selected two-parameter correction, nats/image. Mean $\pm$ sample SD over three seeds; $\Delta$ is output minus path. Each run is stopped by its own validation criterion; Caps counts the path/output runs that reached the 12,000-update limit.}
\label{tab:heterogeneous-training}
\begin{tabular}{llrrrr}
\toprule
Data & Model & Path & Output & $\Delta$ & Caps\\
\midrule
Digits & MLP & $18.091\pm0.052$ & $17.771\pm0.039$ & $-0.319\pm0.027$ & 0/0\\
Digits & Pair & $20.194\pm0.025$ & $19.755\pm0.004$ & $-0.439\pm0.029$ & 0/0\\
MNIST & MLP & $110.402\pm0.164$ & $108.331\pm0.365$ & $-2.071\pm0.468$ & 0/0\\
MNIST & Pair & $119.294\pm0.104$ & $114.446\pm0.275$ & $-4.848\pm0.341$ & 0/3\\
\bottomrule
\end{tabular}
\end{table}

Table~\ref{tab:heterogeneous-training} shows that output training lowers
the calibrated test NLL in every one of the 12 paired comparisons (two data
sets, two models, three seeds), by $0.32$ to $4.85$ nats/image on average,
while raising mean-path NLL: the tradeoff described by
Theorem~\ref{thm:main}. Three MNIST logistic-field output runs reach the
12,000-update limit; the other 21 runs stop when validation no longer
improves.

\paragraph{The advantage survives stronger baselines.}
Three controls strengthen the path-trained baseline. First,
\emph{pair-specific recalibration}: the explicit pair weights of the
logistic field permit the retention-specific transformation
$(1+\lambda)/(c_{ij}+\lambda)$ of Corollary~\ref{cor:regularized}. Jointly
selecting $\lambda$ and the logit correction on validation reduces the
path-trained NLL to $19.958\pm0.009$ on digits and $118.713\pm0.133$ on
MNIST; the same selection gives $19.755\pm0.004$ and $114.540\pm0.270$
after output training, which remains lower in all three seeds. Second, \emph{matched
budgets}: truncating both objectives to a common training-context budget
preserves the direction in all three seeds of each model and data set
(online appendix, Table~S1), and so does truncating both runs at
a common wall-clock time. Third, \emph{a stronger path optimizer}:
Table~\ref{tab:strong-mlp-controls} adds a cheaper unbiased estimator of
the same path loss, which samples one round per example; its learning rate
is chosen from three values, each run being given the wall-clock time of
the slower of the two main runs. Output training still wins all six data set/seed comparisons,
by $0.326\pm0.094$ and $3.067\pm0.322$ nats on digits and MNIST. A greedy
confidence rule, which reveals in each round the pixels whose logits have
the largest magnitude, applied to the same trained models, scores worse
than the heterogeneous library by 1.23--2.75 nats on digits and
58.43--77.80 on MNIST.
Output training is also inexpensive at this scale: the three output-trained
MLP runs take 30.85\,s on digits and 154.89\,s on MNIST including start-up,
whereas the learning-rate search of the stronger baseline uses 133.29\,s
and 501.04\,s (Appendix~\ref{app:extension-protocol}).

\begin{table}[t]
\centering\small
\caption{Stronger MLP controls in the same heterogeneous library after
validation-selected two-parameter correction, nats/image
(three-seed mean $\pm$ sample SD).}
\label{tab:strong-mlp-controls}
\begin{tabular}{lrrr}
\toprule
Data & Full path & Sampled path & Output\\
\midrule
Digits & $18.091\pm0.052$ & $18.097\pm0.079$ & $17.771\pm0.039$\\
MNIST & $110.402\pm0.164$ & $111.398\pm0.076$ & $108.331\pm0.365$\\
\bottomrule
\end{tabular}
\end{table}

\subsection{Retention diagnostics and library design}
\label{sec:diagnostics}
Squared training-pixel correlations give a pair-energy proxy from which
the residual $R$, the correction cost $B$ and the heterogeneity gap $H$ of
any bank can be computed without using validation or test data
(Appendix~\ref{app:retention-proxy}).
Table~\ref{tab:retention-diagnostics} reports them for the banks used in
the experiments.

\begin{table}[t]
\centering\small
\setlength{\tabcolsep}{4pt}
\caption{Pair-energy proxies computed from training data only, and test
NLL of the evaluated banks. $R,B,H$ are normalized proxies of the residual,
correction cost and heterogeneity gap; $\E R_K$ is
\eqref{eq:finite-bank-residual}. NLL is nats/image after two-parameter
correction. Uniform rows average two separate banks; their $M=1$ models are
held fixed across $K$ and differ from the two-block models. A dash marks a
measurement not made.}
\label{tab:retention-diagnostics}
\begin{tabular}{@{}llrrrrrr@{}}
\toprule
Data / $K$ & Library & $R$ & $\E R_K$ & $B$ & $H$ & Path NLL & Output NLL\\
\midrule
Digits / 10 & Two-block & 0.231 & -- & 0.872 & 0.199 & 18.091 & 17.771\\
Digits / 10 & Uniform & 0.078 & 0.075 & 0.129 & 0.030 & 17.078 & --\\
MNIST / 18 & Two-block & 0.274 & -- & 1.185 & 0.219 & 110.402 & 108.331\\
MNIST / 18 & Uniform & 0.073 & 0.072 & 0.110 & 0.018 & 97.704 & --\\
MNIST / 128 & Uniform & 0.063 & 0.063 & 0.086 & 0.003 & 94.786 & --\\
MNIST / 512 & Uniform & 0.062 & 0.062 & 0.084 & 0.001 & 93.150 & --\\
\bottomrule
\end{tabular}
\end{table}

\paragraph{Finite banks behave as predicted.}
The expectation~\eqref{eq:finite-bank-residual} holds for any fixed proxy
weights. On digits at $K=10$ it is $0.07483$, against $0.07475$ and $0.08045$
on the two banks; on MNIST at $K=18/128/512$ it is $0.07241/0.06348/0.06239$,
against observed bank means $0.07260/0.06349/0.06244$.

\paragraph{Library choice and training objective.}
The two-block library was built to create heterogeneity, and its gap $H$
is accordingly about seven times (digits) and twelve times (MNIST) that of
the same-size uniform banks. It pays for this with three to four times the
residual and seven to eleven times the correction cost. Uniform banks with
their own path-trained models reach $17.078$ and $97.704$ nats/image, below
even the output-trained two-block models ($17.771$ and $108.331$), and
enlarging the uniform MNIST bank from $K=18$ to $128$ to $512$ with fixed
models lowers NLL from $97.704$ to $94.786$ to $93.150$. The two decisions
play different roles: the library sets how much dependence survives, and
the training objective determines how much of the lost dependence is
recovered within that library. The online appendix also reports the
moments of each bank and comparisons between banks of the same size.

\subsection{Output training absorbs the pair correction}
\label{sec:unit-gain}
At the population output optimum every covered pair is already restored,
so the context gain selected after output training should be one. With
the marginal coefficient fixed at $\alpha=1$, the logistic field selects mean
gains of $1.225$ after path training and $1.008$ after output training on
digits, and $1.100$ and $1.025$ on MNIST: output training absorbs the
correction that the path-trained model needs from calibration and ends at
the predicted unit gain. The MLP moves in the same direction, from $1.033$
to $0.800$ on digits and from $1.100$ to $0.967$ on MNIST. The online
appendix lists the selections for all 24 checkpoints.

\subsection{Serial normalization in a frozen language model}
\label{sec:text}
We apply context correction to the frozen 169.6-million-parameter masked
diffusion language model (MDLM) of \citet{mdlm}, trained on OpenWebText, with uniform fixed-quota libraries at
$T=8$ and $16$ rounds. Thirty-two WikiText-2 validation windows select
corrections separately for each of two $K=64$ banks, and 64 consecutive
test windows of length 128, with no observed prefix, are scored exactly
under each bank (Appendix~\ref{app:text-protocol}).

\begin{table}[t]
\centering\small
\caption{Frozen MDLM NLL after correction selected on the same bank,
nats per 128-token window. Entries average scores from two separate
$K=64$ banks, with independent validation selection. The column $1/c_2$
applies the predicted gain for uniform fixed-quota batching.}
\label{tab:text-finite}
\begin{tabular}{rccccc}
\toprule
$T$ & Identity & Gain $1/c_2$ & Tuned gain & Temperature & Two parameters\\
\midrule
8 & 591.085 & 597.005 & 591.031 & 590.404 & 590.532\\
16 & 572.375 & 574.793 & 571.761 & 571.243 & 571.084\\
\bottomrule
\end{tabular}
\end{table}

Validated corrections improve the frozen model at both round counts
(Table~\ref{tab:text-finite}); the two-parameter correction improves both
banks at both $T$. Applied directly to the raw logits, the predicted gain
$1/c_2$ overshoots, raising NLL by $5.921$ and $2.417$ nats/window, and
validation selects gains slightly below one ($0.95$ for both $T=8$ banks
and $0.95/0.925$ at $T=16$).

A learned predictor can carry a calibration error common to parallel and
serial revealing, and~\eqref{eq:serial-normalization} removes such a
common factor. Serial $T=128$ validation on two fixed orders selects gains
of $0.900$ and $0.875$, so the frozen model's contextual logits call for
shrinkage even when every coordinate is revealed alone. Dividing the mean
gain selected on the two banks by the mean serial gain gives $1.070$ at
$T=8$ and $1.056$ at $T=16$, compared with the predicted values
$1/c_2=1.134$ and $1.058$; all bank/order cross-pairings give
$1.056$--$1.086$ and $1.028$--$1.086$, respectively. At $T=16$ the
normalized gain matches $1/c_2$ to within the resolution of the $0.025$
gain grid. For a finite bank, \eqref{eq:serial-normalization} replaces
$1/c_2$ by the energy-weighted ratio $S_1/S_2$, which tends to $1/c_2$ as
$K$ grows.

\FloatBarrier
\section{Discussion}\label{sec:discussion}
\paragraph{What the spectrum explains.}
A single spectrum, the probability that an interaction has a unique latest-revealed coordinate, governs the local behavior of randomized parallel sampling. It determines how much dependence a schedule mixture retains, what restoring it costs in conditional loss, how finite-mixture likelihood training responds, and how much of that response a single scalar gain can reproduce. Controlled experiments test these consequences on trained predictors: output-likelihood training wins every paired comparison within a heterogeneous library and survives every calibration and optimizer control, the pairwise field ends output training at the predicted unit gain, finite schedule banks match their expected residuals, and a frozen language model's serially normalized gains lie between one and $1/c_2$, matching it at $16$ rounds.

\paragraph{Practical guidance.}
Three rules of thumb follow. First, for uniform fixed-quota batching, a natural starting point for a post-hoc context gain is the model's own serial gain multiplied by $1/c_2$ from \eqref{eq:quota-c2}. Second, output-likelihood or $M$-sample training is worth its extra cost mainly for libraries with heterogeneous retention, such as semi-autoregressive block decoding: by Corollary~\ref{cor:postscalar} and \eqref{eq:heterogeneity-limit}, its advantage over scalar calibration vanishes under homogeneous retention and is at most the heterogeneity gap $H$ otherwise, and the experiments show the same contrast. Third, within a supplied library, the convex calibration-cost criterion of Appendix~\ref{app:spectral-applications} gives a principled objective for schedule weights, and large random banks approach homogeneous retention, with residual $(1-c_2)^2+c_2(1-c_2)/K$ in expectation.

\paragraph{Limitations and open problems.}
The theory is local around a product law and concerns a shared scalar kernel, and trained networks need not reach its population optimum: the MLP ends output training below unit gain, and the finite-$M$ residual predicted for $M=4$ is not visible in the MNIST serial ratios (Appendix~\ref{app:extension-protocol}). Extending the characterization beyond weak dependence, learning schedule libraries from data, training language models with output-likelihood objectives and estimating interaction-specific corrections at scale are natural next steps.

% Funding (last 36 months) and competing interests, as JMLR requires, and the use of AI tools.
\acks{This work received no third-party funding or support, and the authors have no competing interests to declare. Part of the theoretical derivations was carried out with Codex (OpenAI), and Claude (Anthropic) was used to polish the writing; the authors take full responsibility for the content.}

\appendix
\section{Notation}
\label{app:notation}
Table~\ref{tab:notation} lists the main symbols with the place where each is defined.

\begin{table}[ht]
\centering\small
\caption{Main notation.}
\label{tab:notation}
\begin{tabular}{@{}>{\raggedright\arraybackslash}p{.27\textwidth}>{\raggedright\arraybackslash}p{.52\textwidth}l@{}}
\toprule
Symbol & Meaning & Defined in\\
\midrule
$s$, $B_{s,t}$, $C_{s,t}$, $T$ & schedule, its batches, coordinates revealed before round $t$, round budget & Section~\ref{sec:setup}\\
$\pi$, $q_{s,K}$, $q_{\pi,K}$ & schedule weights, component law, output law & \eqref{eq:law}\\
$K_j(\cdot\mid x_C,C)$, $K^{\rm or}$ & shared scalar kernel; exact target conditionals & Section~\ref{sec:setup}\\
$\calC_p$, $\calL_{\rm path}$, $\calL_{\rm out}$, $\calL_M$ & calibration loss; path, output and $M$-sample objectives & \eqref{eq:calibration}, \eqref{eq:train}, \eqref{eq:finiteM-loss}\\
$\mu$, $p_\epsilon$, $f$ & product reference, target family, first-order score & \eqref{eq:target}\\
$f_A$, $P_A$ & interaction components and their projections & Section~\ref{sec:spectrum}\\
$g_{j,C}$, $\alpha_{j,C}$, $\norm{\cdot}_\pi$ & kernel scores, visitation probabilities, calibration norm & \eqref{eq:score}, \eqref{eq:norm}\\
$L_\pi$, $L_\pi^*$, $g^{\rm or}$ & response map, its adjoint, oracle score & \eqref{eq:L}, \eqref{eq:spectrum}\\
$c_A$, $R_\pi$ & retention of $A$, retention operator & \eqref{eq:retention}\\
$c_k$, $c_2$ & retention of order-$k$ interactions and of pairs under uniform batching & \eqref{eq:quota-ck}, \eqref{eq:quota-c2}\\
$B$, $B_*$, $\lambda$ & calibration budget, budget of full recovery, regularization weight & \eqref{eq:frontier}, \eqref{eq:bstar}, \eqref{eq:ridge}\\
$\gamma$, $\gamma^*_{\rm loc}$, $\gamma_{\rm residual}(M)$ & context gain, best local scalar, residual gain after $M$-sample training & \eqref{eq:context-gain}, \eqref{eq:weighted-scalar-gain}, \eqref{eq:finiteM-residual-gain}\\
$E_A$, $E_0$, $S_r$, $a_M(c)$ & interaction energy, uncovered energy, retention moments, finite-$M$ factor & Section~\ref{sec:calibration-geometry}\\
$D_{\rm scalar}$, $D_{M,\rm scaled}$, $H$ & best-scalar risk, risk after training and scaling, heterogeneity gap & \eqref{eq:best-scalar}, \eqref{eq:postscalar}, \eqref{eq:heterogeneity-budget}\\
$R_K$, $w_{ij}$, $\widehat c_{ij}$ & residual of a bank, normalized pair energies, empirical retentions & \eqref{eq:finite-bank-residual}\\
$\gamma_T^*$, $\gamma_0^*$, $\beta$ & parallel and serial gains, common miscalibration factor & \eqref{eq:serial-normalization}\\
$\alpha$, $\ell_\varnothing$, $\ell_C$ & two-parameter correction: marginal coefficient, all-masked and contextual logits & Section~\ref{sec:calibration-geometry}\\
\bottomrule
\end{tabular}
\end{table}

% The notation table may float to the top of the next page; the next section
% heading should not force it out (placeins), so the barrier is skipped once.
\makeatletter\let\retention@FloatBarrier\FloatBarrier
\def\FloatBarrier{\global\let\FloatBarrier\retention@FloatBarrier}\makeatother
\section{Proof of Theorem~\ref{thm:main}}
\label{app:main-theorem-proof}
\paragraph{The interaction decomposition.}
For mean-zero $f$, inclusion--exclusion gives
\[
 f_A=\sum_{C\subseteq A}(-1)^{|A|-|C|}\E_\mu[f\mid X_C].
\]
Integrating out one coordinate of $A$ pairs and cancels terms, proving the coordinatewise mean-zero property. Distinct interaction spaces are orthogonal because a coordinate in their symmetric difference can be integrated out. Conditional expectation onto $X_C$ keeps exactly the interactions contained in $C$.

\paragraph{Adjoint and spectrum.}
For every score field, conditioning and its zero $\mu_j$-mean give
\[
 \inner{f}{L_\pi g}_{L^2(\mu)}
 =\sum_{j,C}\alpha_{j,C}\inner{\E[f\mid X_{C\cup\{j\}}]-\E[f\mid X_C]}{g_{j,C}}_{L^2(\mu)}.
\]
This proves the adjoint formula. The context projection retains $f_A$ precisely when $j\in A$ and $A\setminus\{j\}\subseteq C$. Along a schedule this happens once if $A$ has a unique latest coordinate, and never otherwise. Summing visitation weights yields $L_\pi L_\pi^*f=\sum_Ac_Af_A$. Equivalently, the nonzero singular values of $L_\pi$ are $\sqrt{c_A}$, with each interaction space's multiplicity.

\paragraph{Oracle and risk coefficients.}
Marginalizing \eqref{eq:target} to $C\cup\{j\}$ and to $C$, then dividing, gives
\[
 \frac{p_\epsilon(x_j\mid x_C)}{\mu_j(x_j)}
 =1+\epsilon\bigl(\E[f\mid X_{C\cup\{j\}}]-\E[f\mid X_C]\bigr)+O(\epsilon^2).
\]
The oracle score is therefore $L_\pi^*f$. Differentiating the path product gives \eqref{eq:L}; linearity of the mixture preserves it. For normalized positive families with density derivatives $a,b$ at $\mu$, the quadratic KL coefficient is $\norm{a-b}^2/2$. This follows by differentiating the finite sum defining KL; linear and normalization terms cancel. Applying it to the output gives \eqref{eq:outputrisk}. Applying it to each conditional and using that $p_{\epsilon,C}\to\mu_C$ gives \eqref{eq:calrisk}.

\paragraph{Minimum-norm lifting.}
Put $h=g-g^{\rm or}$ and $r=(I-R_\pi)f$. Then the output residual is $r-L_\pi h$. The range of $L_\pi$ is the orthogonal sum of the spaces with $c_A>0$, since $L_\pi L_\pi^*=R_\pi$. For any $u=\sum_{c_A>0}u_A$ in this range, its minimum-norm preimage is
\begin{equation}
 h_u=L_\pi^*R_\pi^+u,\qquad
 \norm{h_u}_\pi^2=\inner{R_\pi^+u}{u}
 =\sum_{c_A>0}\frac{\norm{u_A}^2}{c_A}.             \label{eq:lifting}
\end{equation}
Indeed, $L_\pi h_u=u$, and every other preimage differs by an element of $\ker L_\pi$, orthogonal to $h_u$. This proves the frontier and its attainability. Taking $u_A=(1-c_A)f_A$ on covered interactions proves \eqref{eq:bstar}; uncovered interactions are necessarily left unchanged.

The resulting total score is
\begin{equation}
 g^*_{j,C}=\sum_{\substack{A\ni j,\ A\setminus\{j\}\subseteq C\\c_A>0}}
                   \frac{f_A}{c_A}.                \label{eq:gstar}
\end{equation}
Exponential normalization makes a positive smooth kernel with this score. If all active interactions are covered, target and output densities agree to first order, so their difference is $O(\epsilon^2)$. Full support bounds $q_{\pi,K}/\mu$ away from zero. The inequality
$\KL(p\|q)\le\sum_x(p(x)-q(x))^2/q(x)$ then yields $O(\epsilon^4)$, completing the proof.\qed

\paragraph{Bounded-budget kernel sequences.}
Let $F_\pi(B;f)$ denote the right-hand side of~\eqref{eq:frontier}.
For any sequence of positive shared kernels, differentiability can be omitted
under a fixed finite calibration budget:
\begin{equation}
 \limsup_{\epsilon\to0}\frac{2\calC_{p_\epsilon}(K^\epsilon)}{\epsilon^2}\le B<\infty
 \quad\Longrightarrow\quad
 \liminf_{\epsilon\to0}\frac{2\KL(p_\epsilon\|q_{\pi,K^\epsilon})}{\epsilon^2}
 \ge F_\pi(B;f).
 \label{eq:budget-sequences}
\end{equation}
Smooth kernels attain the frontier. Appendix~\ref{app:budget-sequences}
proves this extension for the same fixed finite system.

\section{Graph structure, schedule design, and learning}
\label{app:spectral-applications}
\subsection{Comparison with every deterministic adaptive rule}
Consider the zero-field Ising family
\begin{equation}
 p_\epsilon(x)=Z_\epsilon^{-1}\exp\!\left(\epsilon\sum_{i<j}w_{ij}x_ix_j\right),
 \qquad x\in\{-1,1\}^d.                             \label{eq:ising}
\end{equation}
Its first-order interactions are $w_{ij}x_ix_j$. The main theorem gives
\begin{equation}
 \KL(p_\epsilon\|q_{\pi,\rm or})
 =\frac{\epsilon^2}{2}\sum_{i<j}(1-c_{ij})^2w_{ij}^2+O(|\epsilon|^3). \label{eq:isingmix}
\end{equation}
For the coupling graph on $[d]$, whose edges are the pairs with $w_{ij}\ne0$, define the minimum within-round squared edge weight
\[
 \tau_T(w)=\min_{a:[d]\to[T]}\sum_{i<j}w_{ij}^2\ind\{a(i)=a(j)\}.
\]
Empty round classes are allowed.
\begin{corollary}[Best deterministic adaptive sampler on Ising models]\label{cor:graph}
For \eqref{eq:ising} with at most $T$ rounds,
\begin{equation}
 \inf_{A,K:\,\text{rounds}\le T}\KL(p_\epsilon\|q_{A,K})
 =\frac{\epsilon^2}{2}\tau_T(w)+O(|\epsilon|^3),    \label{eq:graph}
\end{equation}
where $A$ ranges over every deterministic value-adaptive rule and $K$ over arbitrary within-round product kernels.
\end{corollary}
\begin{proof}
For any deterministic reveal tree, factoring the target and sampler along its unique output path gives
\begin{equation}
 \KL(p\|q_{A,K})=\E_p\sum_t\left[\TC_p(X_{B_t}\mid h_t)
 +\sum_{j\in B_t}\KL(p(X_j\mid h_t)\|K_{j,h_t})\right]. \label{eq:adaptive}
\end{equation}
The exact-kernel part is the conditional-total-correlation identity of \citet{wen}. For a fixed observed assignment $h$ and fresh set $B$, the first-order conditional density on $B$ consists of singleton fields and direct pair interactions internal to $B$. Marginalizing other unrevealed spins removes their pair terms at first order. The singleton fields appear in the product of conditional marginals and cancel in its density difference. Orthogonality therefore gives
\[
 \TC_{p_\epsilon}(X_B\mid h)=\frac{\epsilon^2}{2}\sum_{i<j:\,i,j\in B}w_{ij}^2+O(|\epsilon|^3).
\]
There are finitely many $(h,B)$, so the remainder is uniform. Every completed adaptive trajectory assigns each vertex one of at most $T$ reveal times, and hence pays at least $\tau_T$. Sum along the path in \eqref{eq:adaptive} and discard nonnegative kernel errors. A fixed schedule attaining $\tau_T$, with exact marginals, gives the matching upper bound.
\end{proof}

For the unit-weight complete graph $K_4$ and $T=3$, $\tau_3=1$. Choose each of the six vertex pairs with probability $1/6$, reveal that pair first, then reveal the other vertices one at a time. Every edge is internal in exactly one schedule, so $c_{ij}=5/6$. Thus the exact-oracle mixture and the optimal deterministic adaptive rule have leading KL coefficients $1/12$ and $1/2$, respectively. The factor of six arises because the mixture loses each edge in only one of the six schedules and the loss enters squared, so the six edges contribute $6\cdot(1/6)^2=1/6$, whereas every deterministic rule with three rounds must place some edge entirely within one round and so contributes at least $1$. The same calculation applies to larger complete graphs by randomizing balanced partitions.

\subsection{A convex criterion for selecting schedule weights}
The minimum additional calibration coefficient for pairwise recovery is
\begin{equation}
 \Psi(\pi)=\sum_{i<j}w_{ij}^2\frac{(1-c_{ij}(\pi))^2}{c_{ij}(\pi)},
 \qquad c_{ij}(\pi)=\sum_s\pi_s\ind\{t_s(i)\ne t_s(j)\}. \label{eq:design}
\end{equation}
A positive-weight edge with zero retention has infinite cost. Since $(1-c)^2/c=c^{-1}-2+c$ is convex for $c>0$ and retention is linear in $\pi$, minimizing \eqref{eq:design} on the simplex is a convex problem for a supplied library. Zero-weight schedules can be discarded. More generally the same objective with $\norm{f_A}^2$ and the retentions $c_A$ applies to any specified interaction family.

For pair interactions with $d\ge2$ and $T\ge2$, a library that covers every pair can be built without knowledge of the target. Assign distinct length-$k$ base-$T$ codes to the vertices, $k=\lceil\log_Td\rceil$, and let each digit define one schedule. Each pair differs in a digit, so uniform weights give $c_{ij}\ge1/k$. Schedule weights can then be optimized using estimated pair strengths while retaining a small uniform floor if desired.

For positive retentions, the gradient of \eqref{eq:design} is
\[
 \frac{\partial\Psi}{\partial\pi_s}
 =\sum_{i<j}w_{ij}^2\bigl(1-c_{ij}^{-2}\bigr)\ind\{t_s(i)\ne t_s(j)\}.
\]
For any feasible $\pi$, convexity supplies an objective-gap bound
\begin{equation}
 \Psi(\pi)-\min_{\rho\in\Delta}\Psi(\rho)
 \le\inner{\pi}{\nabla\Psi(\pi)}-\min_s[\nabla\Psi(\pi)]_s. \label{eq:gap}
\end{equation}
This gives a checkable stopping criterion for schedule-weight optimization. The criterion optimizes the theorem's calibration cost; it is not a claim that the smallest finite-coupling output KL always has the same optimizer.

\subsection{Interaction correction from samples}
Given $n$ independent observations of \eqref{eq:ising}, estimate all pair moments by
$\widehat m_{ij}=n^{-1}\sum_{a=1}^nX_i^{(a)}X_j^{(a)}$.
The regularized kernel suggested by Corollary~\ref{cor:regularized} is
\begin{equation}
 \widehat K_{j,\lambda}(1\mid x_C,C)
 =\operatorname{sigmoid}\!\left(2\sum_{i\in C}\frac{1+\lambda}{c_{ij}+\lambda}\widehat m_{ij}x_i\right). \label{eq:estimated}
\end{equation}
At $\lambda=0$ this is inverse retention; as $\lambda\to\infty$ it approaches the uncorrected moment field. The estimator is supplied the pairwise zero-field model class and library, but neither graph edges, couplings, normalizing constants, nor conditional labels.

\begin{corollary}[Finite-system sample guarantee]\label{cor:sample}
For a fixed Ising family with $d\ge2$ and a library with $c_{ij}>0$ for every pair, let $m=\binom d2$. There exist $C,\epsilon_0,n_0>0$ such that, for every $\delta\in(0,1)$, $|\epsilon|\le\epsilon_0$ and $n\ge n_0\log(2m/\delta)$, the unregularized estimator satisfies
\begin{equation}
 \Prb\!\left\{\KL(p_\epsilon\|q_{\pi,\widehat K_{0}})
 \le C\left[\epsilon^4+\frac{m\log(2m/\delta)}n\right]\right\}\ge1-\delta. \label{eq:sample}
\end{equation}
\end{corollary}
The proof is in Appendix~\ref{app:sample}. Small retention raises the required kernel-field magnitudes and affects the finite-system constants. It does not by itself imply a $1/c_{ij}^2$ leading output-variance penalty: the forward response multiplies the inverse-corrected score by $c_{ij}$ again. This distinction follows from the same operator identity.

\section{Numerical verification}
\label{app:numerical-verification}
The computations check the response spectrum and the frontier of Theorem~\ref{thm:main}, the comparison with deterministic adaptive samplers in Corollary~\ref{cor:graph}, and the correction estimated from samples in Corollary~\ref{cor:sample}. All finite-state calculations use full sums in float64.

\subsection{Response spectrum and the sharp frontier}
An independent implementation constructs $L_\pi$ from raw state functions and context visitation weights. We use a biased four-binary-variable product law and a product alphabet of sizes $2\times3\times2$, with nonuniform coordinate masses and nonuniform schedule weights. An orthonormal product basis is formed by weighted Gram--Schmidt. Table~\ref{tab:full-state-checks} compares directly assembled Gram matrices with the retention spectrum. Minimum-norm solutions are checked against a numerical pseudoinverse, and the regularized solutions against direct normal-equation solves.
\begin{table}[ht]
\centering\small
\begin{tabular}{lrrr}\toprule
Reference & States & $\max|LL^*-R|$ & Density-derivative error\\\midrule
Biased binary, $d=4$ &16&$8.9\times10^{-16}$&$4.3\times10^{-11}$\\
Mixed alphabet $2\times3\times2$&12&$6.7\times10^{-16}$&$5.9\times10^{-11}$\\\bottomrule
\end{tabular}
\caption{Full-state checks. Density derivatives are central differences at $\epsilon=10^{-5}$ using normalized shared kernels.}
\label{tab:full-state-checks}
\end{table}
The output--calibration objective agrees with \eqref{eq:ridgevalue} to numerical precision for $\lambda\in\{0.01,0.1,1,10\}$, including libraries with uncovered interactions. Figure~\ref{fig:calibration-frontier} displays the spectral frontier against the numerical solutions. Direct finite-density evaluations also check both KL expansions and the bound in Proposition~\ref{prop:finite}. These checks separately exercise categorical alphabets, shared contexts, nonuniform priors, and uncovered interactions.

\subsection{Ising graphs, learning from samples and schedule design}
We solved the complete Bellman recursion over deterministic adaptive policies for $K_4$ with three rounds, a five-cycle with two rounds, and $K_5$ with three rounds. At $\epsilon=10^{-4}$ their optimal KL divided by $\epsilon^2$ is $0.5000000$, $0.5000000$, and $1.0002001$, approaching the coefficients $1/2,1/2,1$ in Corollary~\ref{cor:graph}. The $K_4$ oracle mixture has ratio approximately $1/5.99494$ to the deterministic optimum.

We also evaluated eight-spin paths, cycles, and a $2\times4$ grid under three two-round schedules defined by the bits of the vertex index, fixed independently of the graph. Oracle conditionals, the best global temperature selected using the exact target, and inverse-retention fields were evaluated for six coupling strengths from $0.0025$ to $0.08$. Table~\ref{tab:slopes} gives the fitted small-coupling slopes.
\begin{table}[ht]
\centering\small
\begin{tabular}{lrrr}\toprule
Graph & Exact conditionals & Best global temperature & Inverse retention\\\midrule
Path&1.9999&2.0004&3.9974\\
Cycle&1.9999&2.0000&3.9974\\
$2\times4$ grid&2.0000&2.0047&3.9958\\\bottomrule
\end{tabular}
\caption{Slopes of log KL against log coupling over the four smallest couplings. The predicted orders are two, two, and four.}\label{tab:slopes}
\end{table}

For learning from samples, all $28$ pair moments are estimated on the grid. At $\epsilon=0.04$, inverse-retention mean KL over three sample streams is $0.001034$ at $16{,}384$ observations and $0.000148$ at $262{,}144$, compared with $0.003558$ and $0.003068$ for the corresponding uncorrected moments. The online appendix reports every sampled training size, including the small-sample regime; no oracle labels are used in training.

Finally, optimizing \eqref{eq:design} over the fifteen nontrivial unoriented cuts of a weighted five-vertex graph reduces the doubled minimum calibration coefficient from $2.0794375$ under uniform weights to $0.6775756$. The numerical first-order gap in \eqref{eq:gap} is $9.8\times10^{-8}$. This directly tests the design objective implied by the theorem.

\section{Displacement of the optimal context gain}
\label{app:gain-displacement}
Even for a weak pairwise Gibbs target and uniform balanced batches, the gain that minimizes output KL can vary linearly with the coupling. Consider four spins and
\begin{equation}
 p_\epsilon(x)\propto
 \exp\!\bigl(\epsilon(x_1x_2+x_1x_3+x_2x_3)\bigr),
 \qquad x\in\{-1,1\}^4,
 \label{eq:triangle-displacement-target}
\end{equation}
with all six ordered $(2,2)$ partitions equally likely. The fourth spin is independent. Here $c_2=2/3$, and the common target correlation among the first three spins is
\[
 t=\frac{e^{4\epsilon}-1}{e^{4\epsilon}+3}.
\]
Apply the exact-oracle gain~\eqref{eq:context-gain} and put
$u=\tanh(\gamma\operatorname{atanh}t)$ and
$v=\tanh(2\gamma\epsilon)$. Half the partitions reveal one triangle vertex in the first batch; their average triangle correlation is $(2u+u^2)/3$. The other half reveal two triangle vertices first, with average correlation $v/3$. The final mixture therefore has common correlation
\begin{equation}
 r_\gamma=\frac{2u+u^2+v}{6}.
 \label{eq:triangle-gain-correlation}
\end{equation}
Both triangle laws are invariant under permutations and global spin flip, so $r_\gamma=t$ is equivalent to equality of their full laws. After dividing $r_\gamma-t$ by $\epsilon$, the implicit function theorem gives a unique root near $\gamma=3/2$. Write $\gamma=3/2+a\epsilon+O(\epsilon^2)$. Since $t=\epsilon+\epsilon^2+O(\epsilon^3)$,
\[
 2u+u^2+v
 =6\epsilon+\left(\frac{21}{4}+4a\right)\epsilon^2
       +O(\epsilon^3),
 \qquad
 6t=6\epsilon+6\epsilon^2+O(\epsilon^3).
\]
Thus the zero-KL, and hence globally optimal, gain satisfies
\begin{equation}
 \gamma^*(\epsilon)=\frac32+\frac{3}{16}\epsilon+O(\epsilon^2).
 \label{eq:triangle-gain-shift}
\end{equation}
The linear coefficient depends on higher-order terms of the target and kernel family.

\section{Positive kernel sequences with bounded calibration}
\label{app:budget-sequences}
We prove~\eqref{eq:budget-sequences}. The target expansion and full support imply that, at every visited context assignment, both its probability and its conditional probabilities are bounded below near zero. There are finitely many such assignments, each with fixed positive visitation weight. A bound $\calC_{p_\epsilon}(K^\epsilon)=O(\epsilon^2)$ therefore bounds each conditional KL by $O(\epsilon^2)$. Pinsker's inequality, followed by the target's conditional expansion, gives $K_j^\epsilon(\cdot\mid x_C,C)-\mu_j=O(|\epsilon|)$ uniformly. Consequently the finite score vector
\[
 (g_\epsilon)_{j,C}
 =\frac{K_j^\epsilon(\cdot\mid x_C,C)/\mu_j-1}{\epsilon}
\]
is bounded and has zero $\mu_j$-mean at each history. Expanding the finite path products and conditional divergences uniformly over this bounded set gives
\begin{align*}
 \frac{2\KL(p_\epsilon\|q_{\pi,K^\epsilon})}{\epsilon^2}
 &=\norm{f-L_\pi g_\epsilon}^2+o(1),\\
 \frac{2\calC_{p_\epsilon}(K^\epsilon)}{\epsilon^2}
 &=\norm{g_\epsilon-g^{\rm or}}_\pi^2+o(1).
\end{align*}
Choose a subsequence attaining the output liminf, then a convergent subsequence of $g_\epsilon$. Its limit $g$ has calibration norm at most $B$, and continuity together with the minimum-norm lifting argument in Theorem~\ref{thm:main} give $\norm{f-L_\pi g}^2\ge F_\pi(B;f)$. Conversely, exponentially normalized fixed score fields attain every frontier point.

\section{Finite-mixture objectives and local response}
\label{app:finiteM}
\begin{lemma}[Expected finite-mixture likelihood]\label{lem:finiteM-Jensen}
For a finite schedule law and positive component distributions,
\eqref{eq:finiteM-monotonic} holds.
\end{lemma}
\begin{proof}
Fix $x$ and let $Y_i=q_{S_i,K}(x)>0$ be independent identically distributed variables. Jensen gives
$\E[-\log(M^{-1}\sum_iY_i)]\ge-\log\E Y_1$.
For $M+1$ samples, define $A_i=M^{-1}\sum_{j\ne i}Y_j$, whose mean is the mean of all $M+1$ samples. Convexity gives
\[
 -\log\!\left(\frac1{M+1}\sum_{i=1}^{M+1}Y_i\right)
 \le\frac1{M+1}\sum_{i=1}^{M+1}[-\log A_i].
\]
Taking expectations proves monotonicity. The strong law gives convergence of the sample mean to $\E Y_1=q_{\pi,K}(x)$; finiteness and positivity give $\min_s q_{s,K}(x)>0$, so the logarithms are bounded and bounded convergence applies. Finally average over the finite target space.
\end{proof}

\begin{proof}[Proof of Theorem~\ref{cor:finiteM}]
Let $L_sg$ be the score of a single scheduled output at $\mu$. The scalar fields visited along this schedule are orthogonal in $L^2(\mu)$: for two coordinates in the same batch, integrate out either coordinate, whose field has zero conditional $\mu_j$-mean and does not occur in the other field; for different batches, integrate out the coordinate revealed later, on which earlier fields cannot depend. Thus
\begin{equation}
 \E_s\norm{L_sg}^2=\norm{g}_\pi^2,\qquad
 \E_s L_sg=L_\pi g.
 \label{eq:finiteM-path-isometry}
\end{equation}
For a fixed schedule tuple, the mixture score is $M^{-1}\sum_mL_{S_m}g$, and the finite-state KL expansion and independence give
\begin{align*}
 J_M(g)
 &=\E_{S_{1:M}}\left\|f-\frac1M\sum_mL_{S_m}g\right\|^2\\
 &=\norm{f-L_\pi g}^2
   +\frac1M\E_s\norm{L_sg-L_\pi g}^2,
\end{align*}
which yields~\eqref{eq:finiteM-variance}. For fixed $M$, interchanging the local limit and the schedule expectation is a finite-sum operation.

Write $g^{\rm or}=L_\pi^*f$ and $V_{\rm or}=\sum_A(1-c_A)\norm{f_A}^2$. The identity $L_\pi L_\pi^*=R_\pi$ gives
\begin{equation}
 J_M(g)=\left(1-\frac1M\right)\norm{f-L_\pi g}^2
       +\frac1M\norm{g-g^{\rm or}}_\pi^2
       +\frac1M V_{\rm or}.
 \label{eq:finiteM-frontier-decomposition}
\end{equation}
The last term is independent of the kernels. Equation~\eqref{eq:pathregret} shows that the additional $\lambda\calL_{\rm path}$ contributes $\lambda\norm{g-g^{\rm or}}_\pi^2$ to the doubled local coefficient, up to another kernel-independent term. For $M>1$, divide by $1-1/M$ and apply the regularized frontier with~\eqref{eq:finiteM-effective-lambda}; substitution gives~\eqref{eq:finiteM-optimum}. For $M=1$, the only kernel-dependent term is $(1+\lambda)\norm{g-g^{\rm or}}_\pi^2$. An uncovered interaction is orthogonal to the range of $L_\pi$ and remains in the output residual. For pairs with a common retention $c$ and $\lambda=0$, the trained response is $cM/[1+(M-1)c]$ times its target score, whose reciprocal is~\eqref{eq:finiteM-residual-gain}.
\end{proof}

\section{Further remarks on scalar corrections and finite banks}
\label{app:remarks}
\paragraph{Context gain, constructive kernels and temperature.}
Applying~\eqref{eq:context-gain} to an exact oracle extrapolates the
complete finite-$\epsilon$ logit, which generally differs from
exponentially normalizing the fixed first-order fields in the
constructive proof of Theorem~\ref{thm:main}. The two share the response
in~\eqref{eq:gain-response}; the explicit finite-radius constants of
Proposition~\ref{prop:finite} refer to the latter construction. A
nonuniform marginal background also distinguishes context gain from
global temperature. If every all-masked logit is zero, the two operations
coincide exactly. In the two-parameter family
$\alpha\ell_{\varnothing}+\gamma(\ell_C-\ell_{\varnothing})$, a fixed
$\alpha\ne1$ preserves the product reference only when its empty-context
logits vanish.

\paragraph{The raw finite-$M$ optimum.}
Heterogeneity alone does not make~\eqref{eq:raw-finiteM} smaller
than~\eqref{eq:best-scalar}. For example, equal energies at $c=0.50,0.51$
give $D_{\rm scalar}=1/5101$ and $D_{4,\rm raw}=124034/1600225$.
These retentions are realized on pairs $01,02$ by reveal-time vectors
$(0,1,0),(0,0,1),(0,1,1)$ with probabilities $0.49,0.50,0.01$.

\paragraph{Expected heterogeneity gap of random banks.}
In the setting of~\eqref{eq:finite-bank-residual}, let
$Z_K=\sum_{\widehat c_{ij}=0}w_{ij}$,
$U_K=\sum w_{ij}\widehat c_{ij}$ and
$V_K=\sum w_{ij}\widehat c_{ij}^2$. Define
$H_K=1-Z_K-U_K^2/V_K$, with the ratio set to zero when $V_K=0$.
Since $\E U_K=c$, $\E V_K=c^2+c(1-c)/K$ and
$\E Z_K=(1-c)^K$, Cauchy--Schwarz gives
\begin{equation}
 0\le\E H_K\le
 \frac{1-c}{Kc+1-c}-(1-c)^K .
 \label{eq:finite-bank-gap-bound}
\end{equation}
Specifically,
$(\E U_K)^2\le\E(U_K^2/V_K)\E V_K$ proves the result, including
banks with $V_K=0$.
Replacing $U_K,V_K$ by their expectations and neglecting $Z_K$
gives the diagnostic reference
\begin{equation}
 h_K^{\rm mom}=\frac{1-c}{Kc+1-c}.
 \label{eq:finite-bank-gap-reference}
\end{equation}
It approximates a realized $H_K$ when those weighted moments concentrate
and uncovered energy is negligible. It is neither an identity for
$\E H_K$ nor a bound on each realized bank: with energy on a single pair,
$H_K=0$ for every bank, whereas $h_K^{\rm mom}>0$ when $c<1$.

\paragraph{Serial normalization with interaction-dependent factors.}
If both samplers in~\eqref{eq:serial-normalization} share positive
factors $\beta_A$, one per interaction, instead of a common $\beta$, the respective minimizers
are
\[
 \gamma_T^*=\frac{\sum_Ac_A\beta_AE_A}
                    {\sum_Ac_A^2\beta_A^2E_A},\qquad
 \gamma_0^*=\frac{\sum_A\beta_AE_A}{\sum_A\beta_A^2E_A}.
\]
The unmodified $S_1/S_2$ formula need not survive this change. Uniform
retention still gives $\gamma_T^*/\gamma_0^*=1/c$, since then
$D_T(\gamma)=D_0(c\gamma)$. Different learned fields for the two
samplers, or additional interaction orders, need not satisfy this
relation.

\section{Finite-radius proof}\label{app:finite}
For a mean-zero bounded field $v$ with $\norm{v}_\infty\le H$, put $A(t)=\log\E e^{tv}$. Then $A(0)=A'(0)=0$ and $A''(t)$ is a variance, bounded by $H^2$. Hence $0\le A(\epsilon)\le\epsilon^2H^2/2$. If $a=|\epsilon|H\le1/4$, then
\[
 |e^{\epsilon v-A(\epsilon)}-1-\epsilon v|
 \le\frac{a^2}{2}+\frac12(a+a^2/2)^2e^{a+a^2/2}\le2a^2.
\]
This proves $|p_\epsilon/\mu-1-\epsilon f|\le2\epsilon^2F^2$.

For a component sampler, write its log density relative to $\mu$ as $\epsilon U_s-V_s$, where $U_s$ is the sum of its $d$ visited fields. Each conditional normalization has nonnegative log normalizer at most $\epsilon^2G^2/2$, so $|U_s|\le dG$ and $0\le V_s\le d\epsilon^2G^2/2$. Using $d|\epsilon|G\le1/4$ in the preceding exponential bound gives
\[
 |q_{s,K}/\mu-1-\epsilon U_s|\le2d^2\epsilon^2G^2,
 \qquad q_{s,K}/\mu\ge e^{-1/4-1/32}>1/2.
\]
These estimates survive averaging over schedules, and coverage of every active interaction gives $\E_sU_s=f$. Therefore
\[
 |(p_\epsilon-q_{\pi,K})/\mu|\le2\epsilon^2(F^2+d^2G^2).
\]
Applying the chi-square upper bound on KL and $q_{\pi,K}\ge\mu/2$ gives \eqref{eq:finite}.

\section{Sample-estimation proof}\label{app:sample}
Let $\phi_{ij}(x)=x_ix_j$ and $\theta_{ij}=\epsilon w_{ij}$. Under uniform $\mu$, distinct pair functions are orthonormal. Differentiating the Gibbs moment at zero gives $m_{ij}(\epsilon)=\E_{p_\epsilon}\phi_{ij}=\theta_{ij}+O(\epsilon^2)$. Write $Q(b)$ for the output law obtained by inserting the vector $b$ in place of $\widehat m$ in \eqref{eq:estimated} at $\lambda=0$. Its map is analytic near zero, and Theorem~\ref{thm:main} gives
\[
 Q(b)/\mu=1+\sum_{i<j}b_{ij}\phi_{ij}+O(\norm{b}_2^2).
\]
The target has the analogous expansion with $b=\theta$. For $e=\widehat m-m(\epsilon)$ in a sufficiently small fixed ball,
\[
 \norm{(p_\epsilon-Q(\widehat m))/\mu}_\infty
 \le C_1\bigl(\epsilon^2+\norm{e}_2\bigr),\qquad Q(\widehat m)\ge\mu/2.
\]
Thus $\KL(p_\epsilon\|Q(\widehat m))\le C_2(\epsilon^4+\norm{e}_2^2)$. Each observed pair product lies in $[-1,1]$, so Hoeffding's bound and a union bound give $\norm{e}_2^2\le2m\log(2m/\delta)/n$ with probability at least $1-\delta$. Taking $\epsilon_0$ small enough and $n_0$ large enough to stay in the fixed ball proves \eqref{eq:sample}. All constants refer to the stated finite system.

\section{Additional results under uniform retention}
\label{app:uniform-controls}
\paragraph{Exact finite-library scores.}
Table~\ref{tab:learned-finite} evaluates the fine-tuned models under two
fixed libraries with $K=512$ at $T=4$, on the first $128$ test images,
which were used in a pilot study (Appendix~\ref{app:observed-protocol}). We average the two libraries' NLLs, then report the mean and
sample standard deviation across the five training seeds. Each library
defines its own output law.

\begin{table}[htb]
\centering\small
\caption{Exact finite-library output NLL in nats/image, mean $\pm$ training-seed
SD. The column Gain $1/c_2$ applies the predicted gain to both training objectives. Calibration
parameters are selected on validation with independently drawn uniform
schedules (Appendix~\ref{app:observed-protocol}) and are not refitted to
these libraries.}
\label{tab:learned-finite}
\begin{tabular}{llcccc}
\toprule
Data & Training & Identity & Gain $1/c_2$ & Temperature & Tuned gain\\
\midrule
Digits & $M=1$ & $16.673\pm0.051$ & $16.379\pm0.041$ & $16.561\pm0.037$ & $16.313\pm0.039$\\
Digits & $M=4$ & $16.458\pm0.080$ & $16.293\pm0.071$ & $16.419\pm0.075$ & $16.306\pm0.044$\\
MNIST & $M=1$ & $94.756\pm0.294$ & $92.407\pm0.345$ & $92.165\pm0.294$ & $92.407\pm0.345$\\
MNIST & $M=4$ & $94.261\pm0.541$ & $93.103\pm0.595$ & $92.967\pm0.508$ & $93.263\pm0.567$\\
\bottomrule
\end{tabular}
\end{table}

\begin{figure}[htb]
\centering\includegraphics{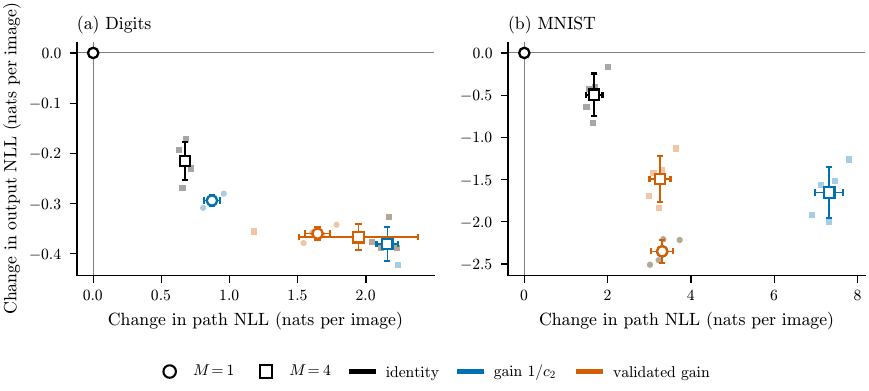}
\caption{Changes in output NLL and path NLL relative to the $M=1$ model with identity calibration, on the same two $K=512$ libraries. Marker shape gives the training objective and color the calibration; small markers are the five training seeds and open markers their mean $\pm$ sample SD. On MNIST, validation selects the gain $1/c_2$ for every $M=1$ seed, so the two calibrated $M=1$ markers coincide. These are empirical loss differences, not a fitted frontier.}
\label{fig:path-output-plane}
\end{figure}

Figure~\ref{fig:path-output-plane} plots the output and path changes
reported in Section~\ref{sec:learned}, together with the calibrated models.

\paragraph{Calibration of the uniform-library models.}
$M=1$ with the gain $1/c_2$ already improves on uncalibrated $M=4$ in every
seed of both data sets, consistent with the partial correction of finite $M$
in~\eqref{eq:finiteM-optimum}. After a validation-selected gain for both
models, $M=4$ changes NLL by $-0.007\pm0.024$ on digits and
$+0.857\pm0.270$ on MNIST relative to $M=1$, and temperature is better than
the gain $1/c_2$ on MNIST. Table~\ref{tab:fine-bank} recalibrates each model
on the two libraries themselves; with two-parameter correction the paired
$M=4$ minus $M=1$ difference is $-0.0006\pm0.0447$ nats/image on digits and
$+1.0347\pm0.2590$ on MNIST for these models. Continued MNIST models,
scored on a separate 512-image range, give $-0.068\pm0.102$
(Appendix~\ref{app:extension-protocol}), which also reports serial
normalization of the $M=4$ models.

\begin{table}[htb]
\centering\small
\caption{Recalibration on the two $K=512$ libraries, nats/image, five-seed mean $\pm$ sample SD on the 128 test images of Table~\ref{tab:learned-finite}. Corrections are selected on validation for each library (step $0.025$ for gain and temperature, $0.05$ for two parameters), and test NLL is averaged across the two libraries.}
\label{tab:fine-bank}
\begin{tabular}{llrrr}
\toprule
Data & Training & Gain & Temperature & Two parameters\\
\midrule
Digits & $M=1$ & $16.304\pm0.037$ & $16.559\pm0.041$ & $16.296\pm0.038$\\
Digits & $M=4$ & $16.287\pm0.070$ & $16.428\pm0.079$ & $16.295\pm0.071$\\
MNIST & $M=1$ & $92.387\pm0.356$ & $92.138\pm0.303$ & $91.357\pm0.339$\\
MNIST & $M=4$ & $93.012\pm0.587$ & $92.750\pm0.494$ & $92.392\pm0.537$\\
\bottomrule
\end{tabular}
\end{table}

\paragraph{Why finite libraries.}
For the full uniform law, Monte Carlo estimates of output NLL converge
slowly: increasing the number of sampled schedules $S$ from $2048$ to
$8192$ at $T=4$ still changes the estimate for the seed-$11$ MNIST MLP by
$-0.420$ nats/image with identity and $-0.814$ with the gain $1/c_2$, and for
the gain-calibrated model the fifth percentile over images of the
effective sample size of the schedule weights is $1.14$. Exact finite-library
scores avoid this integration error. At $K=32,128,512$, the $M=4$ minus
$M=1$ identity NLL differences on MNIST are $+0.021$, $-0.285$, and
$-0.495$; each $K$ defines a different output law.

\paragraph{Trained deterministic and value-adaptive controls.}
On the same 128 images, Table~\ref{tab:fixed-controls} compares the
randomized library with two separately trained fixed schedules, both with
the same training-context budget. Within each row the calibrators have the same one-parameter form;
the digits fixed controls have six gain candidates, while the randomized
library has seven. The likelihoods of the fixed schedules are exact.
The randomized law improves on these two fixed controls in all five seeds.
These are two concrete algorithms, rather than a search over the full
deterministic or adaptive class.

\begin{table}[htb]
\centering\small
\caption{Trained controls, nats/image on the same 128 test images,
five-seed mean $\pm$ sample SD. The $M=1$ column averages exact NLLs of
the two $K=512$ libraries; the fixed columns use the exact law of one
random permutation cut into four equal rounds, or of round $(r+c)\bmod4$
for the pixel in row $r$ and column $c$ (diagonal residue).}
\label{tab:fixed-controls}
\begin{tabular}{llrrr}
\toprule
Data & Calibration & Randomized $M=1$ & Fixed random & Diagonal residue\\
\midrule
Digits & Tuned gain & $16.313\pm0.039$ & $18.120\pm0.102$ & $17.996\pm0.086$\\
MNIST & Temperature & $92.165\pm0.294$ & $96.836\pm0.400$ & $95.669\pm0.426$\\
\bottomrule
\end{tabular}
\end{table}

A further control applies a greedy confidence rule to the base
MLP, revealing in each round the coordinates whose raw logits have the
largest magnitude. At seed 11 and $T=4$, its uncalibrated NLL exceeds that
of the same network under the $K=512$ libraries by $3.462$ nats/image on
digits and $71.252$ on MNIST, on the same 128 images; the network is not
retrained for this rule.

\section{Experimental protocols}
\label{app:protocols}
This appendix collects what is needed to read and reproduce
Section~\ref{sec:experiments}. The online appendix gives the complete
protocols, timing measurements, numerical cross-checks, per-bank
diagnostics and the post hoc comparisons between banks.

\subsection{Data, models and uniform-library training}
\label{app:observed-protocol}
The digits are the $1797$ scikit-learn $8\times8$ images, binarized at
intensity $\ge8$ (of $16$) with a fixed stratified $1078/359/360$
train/validation/test split; MNIST is binarized at intensity $\ge128$ (of
$255$) with a $50{,}001/9{,}999/10{,}000$ split. The first $1000$ MNIST test images
were used in a pilot study, and the heterogeneous-library experiments use
test images $1000$ to $1511$ as a new test set. Training and calibration use only
training and validation observations.

The logistic field has a learned bias and a symmetric off-diagonal weight
matrix acting on centered observed pixels, so it implements the pairwise
first-order field. The MLP has two width-$256$ GELU hidden layers and
receives the revealed pixels as $\pm1$ values, with zeros elsewhere, and
visibility indicators. With batch size
$256$, base training uses Adam for $4000$ digits or $6000$ MNIST updates, at
learning rate $0.003$ for the logistic field and $0.001$ for the MLP, from
initialization seeds $11,22,33,44,55$; validation binary cross-entropy
(BCE) on fixed random masks selects the checkpoint. Fine-tuning uses Adam at $0.0003$, the same base
checkpoint per seed, and $T=4$; the $M=1$ and $M=4$ runs use $1950$ and
$600$ updates and process $1{,}996{,}800$ training contexts (masked inputs
passed through the network) per seed.
The two fixed-schedule controls use a random permutation (seed $8122026$)
or the residue of row plus column modulo four, with the same
training-context budget. The confidence control fills each batch with the
coordinates whose raw logits, given the revealed values, have the largest
magnitude, breaking ties by index.

Post-training calibration uses the first $128$ validation images and $128$
independent fixed-quota schedules per image. Temperature logit scales are
$0.4,0.55,0.7,0.85,1,1.15,1.3,1.6$, and gain candidates are
$\{0.5,1,1/c_2,0.7/c_2,0.85/c_2,1.15/c_2,1.3/c_2\}$. Selected parameters are
fixed before test scoring. The finite-library evaluation uses the same
fixed schedule bank for every image, with two banks and $K=32,128,512$,
and sums each bank's density exactly. Mean path-NLL differences between
models on the same bank equal differences in conditional calibration loss,
since the oracle term cancels in~\eqref{eq:pathregret}.

\subsection{Heterogeneous library and controls}
\label{app:extension-protocol}
The heterogeneous library is built as described in
Section~\ref{sec:heterogeneous-learning}; its $K=2b$ schedules have uniform
weights, and on MNIST the two rounds of a half can have unequal sizes.
Both model classes start from the same model-specific random-mask
checkpoint within each seed ($11,22,33$), and every update uses the entire
library. Adam uses learning rate $0.0003$, with batch sizes $128$ for digits
and $64$ for MNIST. Every $100$ updates a checkpoint is scored on the first
$128$ validation images by its lowest exact validation output NLL over
$\alpha\in\{0.75,1,1.25\}$ and $\gamma\in\{0.75,1,1.25,1.5,1.75,2\}$; eight
checks without improvement stop training, and the update limit extends
from $4000$ to $8000$ and then $12000$ only if this criterion has not
stopped the run. The final post-training correction uses the same $128$
validation images, step-$0.025$ grids for gain and temperature, and a step-$0.05$
two-parameter grid over $\alpha\in[0.5,1.75]$ and $\gamma\in[0.5,2.5]$.
Digits uses all $360$ test images; MNIST uses test images $1000$ to $1511$,
fixed before training, and the later controls reuse these images.

Pair-specific correction multiplies each off-diagonal logistic-field weight
by $(1+\lambda)/(c_{ij}+\lambda)$, with
$\lambda\in\{0,0.1,0.25,0.5,1,2,4,\infty\}$ chosen on validation together
with the logit correction. For the matched-budget control, both objectives
are truncated at the smaller of the two training-context counts reached by
their independently stopped runs (online appendix, Table~S1). The
sampled-path baseline of Table~\ref{tab:strong-mlp-controls} samples a schedule and a round with probability
proportional to the round's width and averages BCE over the coordinates
revealed in that round; this estimator and its gradient are unbiased for
mean-path NLL divided by $d$. Each of the learning rates
$10^{-4},3\cdot10^{-4},10^{-3}$ receives the longer of the two main
runs' wall-clock times for the same seed and data set, and only checkpoints
saved within that time are eligible. The three-rate search
consumes $133.29$ seconds on digits and $501.04$ seconds on MNIST across
three seeds; the corresponding full-library output runs take $30.85$ and
$154.89$ seconds, including start-up.

Continuing all ten uniform-library MNIST runs to $8000$ updates, with
preserved optimizer and random-generator states, gives exact $K=512$ NLL
after two-parameter correction of $90.520\pm0.411$ for $M=1$ and
$90.452\pm0.411$ for $M=4$ on MNIST test images $1000$ to $1511$, a paired
difference of $-0.068\pm0.102$; the two objectives consume different numbers
of training contexts. Serial-normalized MNIST ratios are $1.441/1.396$ for
$M=1$ and $1.452/1.406$ for $M=4$ across the two orders; the latter do not
show the smaller residual gain predicted for the local $M=4$ population
optimum.

\subsection{Frozen language model}
\label{app:text-protocol}
The checkpoint \texttt{kuleshov-group/mdlm-owt} has $12$ transformer
layers, $12$ attention heads, hidden width $768$ and $169{,}627{,}218$
parameters. We load the released weights in FP32 into a PyTorch
implementation that uses scaled dot-product attention; as in the released
configuration, time conditioning is disabled, and the mask token is
excluded from the output softmax. The
checkpoint and the GPT-2 tokenizer are pinned at the revisions
\begin{center}\footnotesize
\begin{tabular}{@{}ll@{}}
checkpoint & \texttt{d0958fa851335ece6c15260ce0025f030673c0fb}\\
tokenizer & \texttt{607a30d783dfa663caf39e06633721c8d4cfcd7e}
\end{tabular}
\end{center}
WikiText-2 uses the
official validation and test splits, with EOS after every row of the data set;
consecutive $126$-token segments are wrapped with BOS and EOS into
$128$-token windows, all of whose positions are masked and scored. The
leading $32$ validation and $64$ test windows are fixed before calibration.
Exact finite laws use two fixed banks of $K=64$ schedules shared across
windows, and the two banks' NLLs are averaged. Gain, marginal coefficient and
temperature use $0.025$-step grids; the two-parameter search starts from a
$0.1$-step grid over $[0.5,1.5]^2$ and refines it locally. All corrections
for the four bank and round-count combinations are fixed before test
scoring. Two serial references use $T=128$ with one fixed order each.

\subsection{Retention diagnostics}
\label{app:retention-proxy}
For a weak pairwise perturbation in standardized product coordinates, the
pair correlation is its first-order interaction coefficient, which motivates
an energy proxy. Let $e_{ij}=\widehat\rho_{ij}^{\,2}$ and $E=\sum e_{ij}$ over
the selected pairs. We report
\begin{align*}
\bar c_E&=E^{-1}\sum e_{ij}c_{ij}, & R&=E^{-1}\sum e_{ij}(1-c_{ij})^2,\\
B&=E^{-1}\sum_{c_{ij}>0}e_{ij}(1-c_{ij})^2/c_{ij}, & H&=E^{-1}\left(S_0-S_1^2/S_2\right),
\end{align*}
where $S_r=\sum_{c_{ij}>0}e_{ij}c_{ij}^r$. The computation uses all $1078$
digits training observations and the first $20000$ MNIST training
observations; all $2016$ digits pairs and the $60000$ strongest MNIST pairs
are retained, the latter carrying $96.60\%$ of the total squared-correlation
proxy. No validation or test data are used. All compared banks cover
every positive-energy selected pair, so $B$ is the full pair
correction-cost proxy. The online appendix lists the correction selected for
each of the $24$ heterogeneous-library checkpoints.

\clearpage
\setcounter{section}{0}
\setcounter{table}{0}
\renewcommand{\thesection}{S\arabic{section}}
\renewcommand{\thetable}{S\arabic{table}}
\begin{center}
{\Large\bf Online Appendix}
\end{center}
\noindent This online appendix gives the complete experimental protocols, timing
measurements, numerical cross-checks, per-bank diagnostics and post hoc bank comparisons,
the settings of the finite-state checks, and two further remarks on finite mixtures.
Equation, section, table and appendix numbers without the prefix~S refer to the main
paper.

\setlength{\emergencystretch}{1.5em}
\section{Protocols for the uniform-library experiments}
\label{S-app:observed-protocol}
\paragraph{Data.}
The digits data are the $1797$ images of $8\times8$ handwritten digits
distributed with scikit-learn, binarized at intensity $\ge8$ on their
$0$--$16$ scale, with a fixed stratified $1078/359/360$
train/validation/test split. MNIST is binarized at intensity $\ge128$ with
a $50{,}001/9{,}999/10{,}000$ split. The first $1000$ MNIST test
images were used in a pilot study; the heterogeneous-library experiments
use test images $1000$ to $1511$, and the remaining MNIST test images are
not used. These binarizations and splits are our own, rather than a
standard statically binarized benchmark. Training and calibration use only
training and validation data.

\paragraph{Models and training.}
The logistic field has a learned bias and a symmetric off-diagonal weight
matrix acting on centered observed pixels, so it implements the pairwise
first-order field; at finite coupling, marginalizing an Ising model can
produce further terms that it does not represent. The MLP has two
width-$256$ GELU hidden layers and receives the revealed pixels as $\pm1$
values, with zeros elsewhere, and visibility indicators. The five initialization seeds are $11,22,33,44$ and
$55$. With batch size $256$, base training uses Adam for $4000$ (digits) or
$6000$ (MNIST) updates, at learning rate $0.003$ for the logistic field and
$0.001$ for the MLP, and validation binary cross-entropy (BCE) on fixed
random masks selects the checkpoint. This criterion weights histories
differently from the path loss of a fixed number of rounds.

Fine-tuning uses Adam at learning rate $0.0003$ from the same base
checkpoint for each seed, with $T=4$. The $M=1$ and $M=4$ runs use $1950$
and $600$ updates, respectively, so that both process $1{,}996{,}800$
training contexts per seed. Validation uses mixtures of $32$ schedules on
up to $512$ leading validation images, at seven checkpoints including
initialization. The two fixed-schedule controls use a random permutation
(seed $8122026$) or the residue of row plus column modulo four, with the
same training-context budget; their validation uses the single fixed
schedule. Optimizer and random-generator states are saved, so that runs can
be continued.

\paragraph{Calibration and evaluation.}
Post-training calibration uses the first $128$ validation images and $128$
independent fixed-quota schedules per image. The temperature candidates,
which multiply the logits, are $0.4,0.55,0.7,0.85,1,1.15,1.3$ and $1.6$; the
gain candidates are $\{0.5,1,1/c_2,0.7/c_2,0.85/c_2,1.15/c_2,1.3/c_2\}$,
without duplicates. Selected parameters are fixed before test scoring.

Monte Carlo estimates under the uniform law draw independent schedules for
each image, in two independent streams with nested counts $32,128$ and
$512$; checks of the trained models with $512$, $2048$ and $8192$ schedules
use the first $128$ pilot images and initialization seed $11$. The
finite-library evaluation instead uses the same fixed schedule bank for
every image, with two banks and $K=32,128,512$. It evaluates the seed-$11$
base models at $T=2,4$ and all five $M=1$ and $M=4$ MLP seeds at $T=4$,
summing each bank's density exactly. Calibration parameters selected under
the uniform law are applied to these banks without reselection. Differences
in mean path NLL between models on the same bank equal differences in
conditional calibration loss, since the oracle term cancels
in~\eqref{eq:pathregret}. The fixed-schedule controls have a different
schedule law and are compared through their own exact output NLL.

\paragraph{Further diagnostics and controls.}
Moment diagnostics draw $4096$ samples at $T=2,4,8$ and compare the
covariance of every pair of active pixels, those with training marginal in
$[0.05,0.95]$, with the validation covariance. The confidence control fills
each batch with the coordinates whose raw logits, given the revealed
values, have the largest magnitude, breaking ties by index. Its reveal
order for an image is unique, so its likelihood can be computed directly,
and the same raw ranking is kept when the logits are calibrated.

\paragraph{Hardware and timing.}
All experiments run on one 48-GB RTX 4090 D GPU with Python $3.12.3$,
PyTorch $2.8.0$ (CUDA $12.8$) and NumPy $2.3.2$. Training-loop times
include validation and checkpointing but not model and data loading; the
five-seed mean loop times for $M=1/M=4$ are $16.81/5.81$ seconds on digits
and $19.64/8.36$ seconds on MNIST. We also reran the training command
without test scoring for seeds $11$ and $22$ and measured the full elapsed
time, including interpreter start-up, imports, loading and exit. For
$M=1/M=4$ it is $19.46/10.30$ and $20.14/9.09$ seconds on digits and
$22.93/12.84$ and $23.64/12.42$ seconds on MNIST. These reruns measure
time only; all eight reproduce the original model parameters exactly.

\subsection{Frozen language model}
\label{S-app:text-protocol}
The checkpoint \texttt{kuleshov-group/mdlm-owt}, at revision
\begin{center}\small\texttt{d0958fa851335ece6c15260ce0025f030673c0fb},\end{center}
has 12 transformer layers, 12 attention heads and hidden width 768, with
169,627,218 parameters. We load the released weights in FP32 into a
PyTorch implementation that uses scaled dot-product attention; as in the
released configuration, time conditioning is disabled and the mask token is
excluded from the output softmax. We checked that this implementation
reproduces the outputs and path scores of the released model; it does not
reproduce the speed of the released FlashAttention implementation.

The GPT-2 tokenizer is pinned at revision
\begin{center}\small\texttt{607a30d783dfa663caf39e06633721c8d4cfcd7e}.\end{center}
WikiText-2 uses the official validation and test splits, with EOS after
every row of the data set, including empty rows. Consecutive 126-token segments
are wrapped with BOS and EOS to form 128-token windows, and all 128
positions, including the two boundary tokens, are masked and scored. The
leading 32 validation and 64 test windows are fixed before calibration.
Because these windows are consecutive, we do not report confidence
intervals that assume independent windows. The candidate grids are fixed in
advance, test data are not used for selection, and scoring uses
microbatches of four windows.

Monte Carlo estimates under the uniform law use $T=8,16$ and nested
schedule counts $S=8,32,64$. Exact finite laws use two fixed banks with
$K=8,32,64$, shared across all windows. A first experiment transferred
calibration parameters selected under the uniform law to these banks; the
main text reports a second experiment that selects the parameters directly
on each $K=64$ bank. The two banks' NLLs are averaged rather than pooled
into one larger mixture, so each $K$ specifies a different sampler. No
language-model training is performed.

The second experiment uses $0.025$-step one-dimensional grids for gain,
marginal coefficient and temperature. The two-parameter search over
$\alpha\ell_{\varnothing}+\gamma(\ell_C-\ell_{\varnothing})$, with both
parameters in $[0.5,1.5]$, starts from a $0.1$-step grid, refines it by
$\pm0.1$ at step $0.025$ around the best coarse value, and includes all
one-dimensional candidates; it is therefore not a full fine
two-dimensional grid. All corrections for the four bank and round-count
combinations are fixed before test scoring. The two serial references use
$T=128$ with one fixed order each; they serve to normalize the gains and
are not a four-round baseline.

The mean effective sample size of the schedule weights is $1.17$--$1.31$
for the uncorrected text banks, so the posterior over schedules is
concentrated; the finite-library likelihood is nevertheless an exact sum.
The serial gains are $0.900/0.875$, whereas the selected parallel gains are
$0.95/0.95$ for $T=8$ and $0.95/0.925$ for $T=16$, so the ratios of mean
gains are $1.070$ and $1.056$. We report all bank/order cross-pairings
because banks and orders have no natural one-to-one pairing.

\section{Protocols for the heterogeneous-library experiments}
\label{S-app:extension-protocol}
\paragraph{Library and training.}
The heterogeneous library is built as described in
Section~\ref{sec:heterogeneous-learning}, with $K=2b$ schedules of uniform
weight. On MNIST the two rounds of a half can have unequal sizes; the
sampled-path baseline below then samples a round with probability equal
to its size divided by $d$. Both model classes start from the same
model-specific random-mask checkpoint for each seed. The main comparison
has two data sets, two model classes, three seeds ($11,22,33$) and two
objectives, for $24$ training runs. Every update uses the whole library and
computes the empty-context forward pass once. Adam uses learning rate
$0.0003$, with batch sizes $128$ (digits) and $64$ (MNIST). Every $100$
updates, a checkpoint is scored on the first $128$ validation images by its
lowest exact validation output NLL over $\alpha\in\{0.75,1,1.25\}$ and
$\gamma\in\{0.75,1,1.25,1.5,1.75,2\}$. Eight checks without improvement
stop training, and the update limit extends from $4000$ to $8000$ and then
$12000$ only if this criterion has not stopped the run.

\paragraph{Calibration and test data.}
The final post-training correction uses the same $128$ validation images and
the evaluated library. The gain and temperature grids have step $0.025$; the
two-parameter grid has step $0.05$ over $\alpha\in[0.5,1.75]$ and
$\gamma\in[0.5,2.5]$ in $\alpha\ell_{\varnothing}+\gamma(\ell_C-\ell_{\varnothing})$
and includes all one-dimensional candidates. The checkpoint is chosen by the
coarse rule above before this correction, rather than by optimizing the
correction over all saved checkpoints. Digits uses all $360$ test images.
On MNIST, test images $1000$ to $1511$ are fixed before training, and
training uses only training and validation data. The later controls reuse
these images after the main results were known, so we regard them as post
hoc controls.

\paragraph{Pair-specific correction.}
Each off-diagonal logistic-field weight is multiplied by
$(1+\lambda)/(c_{ij}+\lambda)$, with
$\lambda\in\{0,0.1,0.25,0.5,1,2,4,\infty\}$; the value $\lambda=\infty$
leaves the weights unchanged. Validation selects $\lambda$ together with the
logit correction before test evaluation. The selected weights are fixed, so
the correction adds no cost at sampling time.

\paragraph{Matched budgets.}
For the matched-budget control, both objectives are truncated at the
smaller of the two training-context counts reached by the independently
stopped runs, and the checkpoint within this budget is selected by its
validation score. Training contexts, validation contexts and calibration
scoring are counted separately. Wall-clock time is recorded at the end of
each validation step, and truncating both runs at a common wall-clock time
gives the same direction. With matched training contexts, the output minus
path differences are $-0.319/-0.430$ nats on digits and $-1.813/-4.177$ on
MNIST for the MLP/logistic field, and all three seeds agree in direction
(Table~\ref{S-tab:context-controls}).

\begin{table}[t]
\centering
\small
\caption{Matched training-context budgets in the heterogeneous library, with validation-selected two-parameter correction. Exact finite-library test NLL in nats/image, mean $\pm$ sample SD across three training seeds. Both objectives evaluate the full library at every update.}
\label{S-tab:context-controls}
\begin{tabular}{llrrr}
\toprule
Data & Model & Path & Output & Paired difference \\
\midrule
Digits & MLP & $18.091\pm 0.052$ & $17.771\pm 0.039$ & $-0.319\pm 0.027$ \\
Digits & Pair & $20.194\pm 0.025$ & $19.764\pm 0.010$ & $-0.430\pm 0.032$ \\
MNIST & MLP & $110.402\pm 0.164$ & $108.590\pm 0.388$ & $-1.813\pm 0.300$ \\
MNIST & Pair & $119.294\pm 0.104$ & $115.117\pm 0.312$ & $-4.177\pm 0.220$ \\
\bottomrule
\end{tabular}
\end{table}

\paragraph{Sampled-path baseline.}
The sampled-path baseline of Table~\ref{tab:strong-mlp-controls} samples a schedule $S\sim\pi$ and a round $R$
with probability $|B_{S,R}|/d$, and averages BCE over the coordinates
revealed in that round; this estimator and its gradient are unbiased for
mean-path NLL divided by $d$. Its sampling uses a separate random-number
generator, so the minibatch stream is the same for all learning rates.
Each of the rates $10^{-4},3\cdot10^{-4}$ and $10^{-3}$ receives a time
budget equal to the longer of the two primary runs for the same seed and
data set, measured as total process time. Validation runs every $1000$
updates, with patience of eight checks and a limit of $100{,}000$ updates,
and only checkpoints saved within the time budget are eligible. The
validation score selects the checkpoint and the learning rate before the
final correction and test evaluation. The search over the three rates takes
$133.29$ seconds on digits and $501.04$ seconds on MNIST over three seeds,
plus $8.70/10.94$ seconds for the final correction, whereas the
corresponding full-library output runs take $30.85/154.89$ seconds,
including start-up.

\paragraph{Other samplers.}
The greedy confidence rule reveals, in each round, the coordinates whose
raw logits have the largest magnitude given the revealed values, breaking
ties by index; logit correction changes its probabilities but keeps this
ranking. Applied to the trained models without retraining, it is worse
than the heterogeneous library by $1.23$--$2.75$ nats on digits and
$58.43$--$77.80$ on MNIST. Two further uniform banks, drawn with seeds
$51001$ and $151004$, give the $K=10/18$ comparison, which matches the
number of schedules of the heterogeneous library and hence the entropy of
its uniform schedule index; larger banks with $K=128$ and $512$ are
reported separately. Every sampler executes a single four-round schedule.
Banks are calibrated separately, and their NLLs are averaged without
pooling densities.

\paragraph{Numerical cross-checks.}
The forty bank evaluations of the uniform-library study were rescored from the
stored per-image results, including the validation selections, and an
independent implementation that evaluates each round separately recomputes
four trained models on two images. For the heterogeneous study, a separate
CPU implementation that evaluates each path directly, with fresh forward
passes, scores eight seed-$11$ models (both data sets, model classes and
objectives) on their first two test images, with identity and with the
selected two-parameter correction. Output NLL differs by at most
$6.16\cdot10^{-6}$ nats/image, and output NLL, mean-path NLL, effective
sample size and the standard deviation of the log path likelihood all agree
within $1.62\cdot10^{-5}$ in their units, below the tolerance of $10^{-4}$.
For the language model, all $24$ validation selections were recomputed from
the saved candidate scores. These checks confirm the reported numbers; they
do not replace independent retraining.

\paragraph{Continued uniform-library models.}
We continued all ten MNIST runs of the uniform-library study with their
optimizer and random-generator states; every run reaches the $8000$-update
limit before validation stops improving. On test images $1000$ to $1511$,
exact $K=512$ NLL after two-parameter correction is $90.520\pm0.411$ for
$M=1$ and $90.452\pm0.411$ for $M=4$, a paired difference of
$-0.068\pm0.102$ that varies across banks. These continued models are
compared on new test images, whereas the earlier comparison used the first
$128$ pilot images, and the two objectives consume different numbers of
training contexts.

In the fine-grid recalibration of these models on their evaluation banks
(Table~\ref{tab:fine-bank} reports $K=512$), none of the $80$
selections across banks, library sizes and checkpoints chose the marginal
coefficient $\alpha=1$. At $K=512$
the mean selected gains for $M=1/M=4$ are $1.625/1.390$ on digits and
$1.390/1.275$ on MNIST. The serially normalized MNIST ratios are
$1.441/1.396$ for $M=1$ and $1.452/1.406$ for $M=4$ across the two serial
orders, so the $M=4$ models do not show the smaller residual gain predicted
for the local $M=4$ population optimum. Serial references are not available
for the digits models.

\section{Retention diagnostics and selected corrections}
\label{S-app:retention-proxy}
For a weak pairwise perturbation in standardized product coordinates, the
pair correlation is the first-order interaction coefficient, so squared
correlations give an energy proxy; on finite image data this proxy is not
an exact interaction energy. Let $e_{ij}=\widehat\rho_{ij}^{\,2}$ and
$E=\sum e_{ij}$ over the selected pairs. We report
\begin{align*}
\bar c_E&=E^{-1}\sum e_{ij}c_{ij}, & R&=E^{-1}\sum e_{ij}(1-c_{ij})^2,\\
B&=E^{-1}\sum_{c_{ij}>0}e_{ij}(1-c_{ij})^2/c_{ij}, & H&=E^{-1}\left(S_0-S_1^2/S_2\right),
\end{align*}
where $S_r=\sum_{c_{ij}>0}e_{ij}c_{ij}^r$. All compared banks cover every
selected pair with positive energy, so $B$ is the complete pair
correction-cost proxy; with an uncovered pair, full first-order recovery
would be impossible and $B$ would not describe the cost. The computation
uses all 1078 digits training images and the first 20000 MNIST training
images, and pixels with zero variance receive zero correlation. All 2016
digits pairs and the 60000 strongest MNIST pairs are retained, the latter
carrying 96.60\% of the total squared correlation. No validation or test
data are used. These statistics were computed after the training results
were known, so we use them as a diagnostic of the mechanism. In the
two-block banks all pair retentions are positive but some higher-order
retentions vanish, and on MNIST 12 of the 18 schedules have unequal round
sizes. Pair-energy proxies ignore both higher-order dependence and model
error, so they cannot by themselves rank the likelihoods of arbitrary
samplers.

\paragraph{Per-bank moments.}
Table~\ref{S-tab:bankwise-residuals} gives the observed normalized moments
of each bank and their expectation for independently drawn schedules.
Correlations are computed in float64 from the training data; they differ
from an earlier float32 computation by less than $7\cdot10^{-6}$ in every
reported quantity. The same MLP checkpoint is used for all three MNIST bank
sizes, and the 36 model/bank/objective scores were recomputed from the
stored per-image results; no model is retrained. The last column is the
moment reference~\eqref{eq:finite-bank-gap-reference}; at $K=10/18$ it is
0.0303/0.0181 on digits/MNIST, close to the observed $H=0.0300/0.0180$.
\begin{table}[t]
\centering\footnotesize
\setlength{\tabcolsep}{3.8pt}
\caption{Per-bank diagnostics. All selected pairs with positive energy are
covered. The last column is a ratio-of-moments reference, not the expected
or realized gap.}
\label{S-tab:bankwise-residuals}
\begin{tabular}{@{}llrrrrrr@{}}
\toprule
Data / $K$ & Bank & $\bar c_E$ & $R$ & $\E R_K$ & $B$ & $H$ & $h_K^{\rm mom}$\\
\midrule
Digits / 10 & Uniform 0 & 0.76131 & 0.07475 & 0.07483 & 0.12430 & 0.02976 & 0.03030\\
Digits / 10 & Uniform 1 & 0.74916 & 0.08045 & 0.07483 & 0.13416 & 0.03029 & 0.03030\\
Digits / 128 & Uniform 0 & 0.76238 & 0.05791 & 0.05811 & 0.07737 & 0.00247 & 0.00244\\
Digits / 128 & Uniform 1 & 0.75961 & 0.05919 & 0.05811 & 0.07927 & 0.00242 & 0.00244\\
Digits / 512 & Uniform 0 & 0.76297 & 0.05656 & 0.05704 & 0.07449 & 0.00065 & 0.00061\\
Digits / 512 & Uniform 1 & 0.75987 & 0.05806 & 0.05704 & 0.07679 & 0.00068 & 0.00061\\
Digits / 10 & Two-block & 0.64205 & 0.23071 & -- & 0.87221 & 0.19927 & --\\
MNIST / 18 & Uniform 0 & 0.75084 & 0.07249 & 0.07241 & 0.10999 & 0.01813 & 0.01809\\
MNIST / 18 & Uniform 1 & 0.75000 & 0.07271 & 0.07241 & 0.11015 & 0.01783 & 0.01809\\
MNIST / 128 & Uniform 0 & 0.75102 & 0.06348 & 0.06348 & 0.08610 & 0.00262 & 0.00258\\
MNIST / 128 & Uniform 1 & 0.75083 & 0.06351 & 0.06348 & 0.08611 & 0.00253 & 0.00258\\
MNIST / 512 & Uniform 0 & 0.75097 & 0.06239 & 0.06239 & 0.08347 & 0.00066 & 0.00065\\
MNIST / 512 & Uniform 1 & 0.75077 & 0.06248 & 0.06239 & 0.08361 & 0.00066 & 0.00065\\
MNIST / 18 & Two-block & 0.57455 & 0.27358 & -- & 1.18516 & 0.21901 & --\\
\bottomrule
\end{tabular}
\end{table}

\paragraph{Bank orderings after scalar correction.}
By~\eqref{eq:best-scalar}, the best scalar correction of the local oracle
has coefficient $E_0+H$, and $E_0=0$ for the selected pairs in these banks,
so $H$ is the relevant proxy. Table~\ref{S-tab:calibrated-bank-order}
compares the same $M=1$ model on the same test images across the two banks
of each size. The 96 scores involved were checked against the stored
per-image results, and all 96 validation selections for identity, gain,
temperature and two-parameter correction were recomputed; no parameter is
fitted on test data. The four mean orderings after two-parameter correction
agree with $H$, and each holds for all three initialization seeds. The
relative differences $|H_1-H_0|/\max(H_0,H_1)$ range from 0.8\% to 3.7\%.
The oracle residual $R$, which favors bank 0 in every row, agrees with the
NLL ordering only on digits.
\begin{table}[t]
\centering\footnotesize
\setlength{\tabcolsep}{3.2pt}
\caption{Post hoc comparison of two banks with the same model. $H_0,H_1$
use normalized pair-energy proxies computed from training data. The four
$\Delta$ columns give bank 1 minus bank 0 test NLL, averaged over three
initialization seeds; negative values favor bank 1. Units are nats/image.
The last column counts seeds whose NLL ordering agrees with $H$, for
gain-only and two-parameter correction respectively. The oracle residual
$R$ favors bank 0 in every row; its values are in
Table~\ref{S-tab:bankwise-residuals}. Digits banks with $K=128$ and $512$
were not scored on test images.}
\label{S-tab:calibrated-bank-order}
\begin{tabular}{@{}lrrrrrrc@{}}
\toprule
Data / $K$ & $H_0$ & $H_1$ & $\Delta$ Id. & $\Delta$ Gain & $\Delta$ Temp. & $\Delta$ Two & Seeds\\
\midrule
Digits / 10 & 0.029756 & 0.030293 & $+0.1038$ & $+0.1259$ & $+0.1125$ & $+0.1304$ & 3/3; 3/3\\
MNIST / 18 & 0.018131 & 0.017830 & $-1.4440$ & $-1.5372$ & $-1.5301$ & $-1.5510$ & 3/3; 3/3\\
MNIST / 128 & 0.002625 & 0.002528 & $-0.3361$ & $-0.4264$ & $-0.4010$ & $-0.4398$ & 3/3; 3/3\\
MNIST / 512 & 0.000661 & 0.000655 & $-0.0422$ & $-0.0551$ & $-0.0416$ & $-0.0635$ & 2/3; 3/3\\
\bottomrule
\end{tabular}
\end{table}

The gain-only comparison is closer to the scalar family of the theory, but
at MNIST $K=512$, seed 22 has $\Delta\mathrm{NLL}=+0.03348$, opposite to
$H$; its two-parameter difference is $-0.00222$. Temperature also reverses
one seed at that bank size. The uncorrected models already show the same
four mean orderings as the corrected ones, so the agreement is not specific
to calibration. Moreover, the three MNIST sizes are prefixes of the same
two banks and reuse the same models and test images, so these four post hoc
comparisons are not independent trials, and we report the agreement of
signs without a significance test. Confirming that local oracle
coefficients predict the likelihoods of trained models would require
evaluation on new banks.

\paragraph{Why the raw residual need not order calibrated laws.}
Even exact local energies do not supply this ordering. Take the product
reference on three unbiased spins and the score $f=x_0x_1+x_0x_2$.
Library A gives equal weight to the reveal-time vectors $(0,1,0)$ and
$(0,0,1)$, and library B to $(0,1,0)$ and $(0,1,1)$. Their pair retentions
are $(1/2,1/2)$ and $(1,1/2)$. The unnormalized oracle coefficients are
$R_A=1/2>R_B=1/4$, whereas~\eqref{eq:best-scalar} gives the best-scalar
coefficients $D_A=0<D_B=1/5$. The Gibbs target
$p_\epsilon\propto\exp(\epsilon f)$ has full support, so calibration
reverses the order of the raw residuals already in the local model.

\paragraph{Selected corrections for every checkpoint.}
Tables~\ref{S-tab:all-selected-calibration}
and~\ref{S-tab:all-fixed-marginal-gains} list the corrections selected on
validation for all 24 heterogeneous-library checkpoints summarized in
Section~\ref{sec:unit-gain}. All 48 selections were recomputed from the
stored validation grids. A ratio based on the two-parameter $\gamma$ also
changes the marginal coefficient, so it is not the
normalization~\eqref{eq:serial-normalization}, which keeps the marginal
coefficient fixed; serial references are not available for these 24
models.

\begin{table}[t]
\centering\small
\caption{Two-parameter corrections selected on validation for all 24 heterogeneous-library checkpoints. Each cell gives $\alpha/\gamma$; no additional training is performed.}
\label{S-tab:all-selected-calibration}
\begin{tabular}{lllccc}
\toprule
Data & Model & Objective & Seed 11 & Seed 22 & Seed 33\\
\midrule
Digits & Pair & Path & $1.100/1.300$ & $1.100/1.300$ & $1.100/1.300$\\
Digits & Pair & Output & $1.025/1.025$ & $1.050/1.050$ & $1.050/1.050$\\
Digits & MLP & Path & $1.050/1.100$ & $1.000/0.950$ & $1.050/1.050$\\
Digits & MLP & Output & $0.950/0.800$ & $0.900/0.700$ & $0.950/0.850$\\
MNIST & Pair & Path & $1.150/1.200$ & $1.150/1.200$ & $1.150/1.200$\\
MNIST & Pair & Output & $1.050/1.050$ & $1.075/1.075$ & $1.050/1.050$\\
MNIST & MLP & Path & $1.100/1.150$ & $1.125/1.125$ & $1.100/1.200$\\
MNIST & MLP & Output & $1.000/0.975$ & $1.000/0.950$ & $1.000/0.975$\\
\bottomrule
\end{tabular}
\end{table}
\begin{table}[t]
\centering\small
\caption{Context gains selected on validation for the same checkpoints with the marginal coefficient fixed at $\alpha=1$, which separates the context gain from a change of the marginal logits.}
\label{S-tab:all-fixed-marginal-gains}
\begin{tabular}{lllccc}
\toprule
Data & Model & Objective & Seed 11 & Seed 22 & Seed 33\\
\midrule
Digits & Pair & Path & $1.225$ & $1.225$ & $1.225$\\
Digits & Pair & Output & $1.000$ & $1.025$ & $1.000$\\
Digits & MLP & Path & $1.100$ & $0.950$ & $1.050$\\
Digits & MLP & Output & $0.800$ & $0.725$ & $0.875$\\
MNIST & Pair & Path & $1.100$ & $1.100$ & $1.100$\\
MNIST & Pair & Output & $1.025$ & $1.025$ & $1.025$\\
MNIST & MLP & Path & $1.100$ & $1.075$ & $1.125$\\
MNIST & MLP & Output & $0.975$ & $0.950$ & $0.975$\\
\bottomrule
\end{tabular}
\end{table}

For the language model, the serially normalized gain divides the mean of
the two bank gains by the mean of the two serial-order gains. Banks and
orders have no natural pairing, so Section~\ref{sec:text} also reports all
cross-pairings. A selection from a finite grid of candidates is not a
confidence interval for a continuous optimum.

\paragraph{Library size.}
Enlarging a realized finite library changes its law, so its exact
likelihood need not improve monotonically. For independently drawn
components, the expected bound improves with the number of components, as
for the standard finite-mixture objective, but this does not order
individual banks. Drawing a fresh schedule for every sample corresponds to
the full schedule distribution, which a large bank approximates only up to
an integration error that must be assessed separately.

\section{Settings of the numerical checks and the complete sample curve}\label{S-app:protocol}
The binary check uses product marginals $(0.35,0.65)$, $(0.4,0.6)$,
$(0.2,0.8)$ and $(0.55,0.45)$, schedule times $(0,1,1,2)$, $(1,0,1,2)$ and
$(1,1,0,2)$, and schedule weights $(0.2,0.3,0.5)$. The mixed-alphabet check
uses marginals $(0.3,0.7)$, $(0.2,0.3,0.5)$ and $(0.4,0.6)$, schedule times
$(0,1,1)$ and $(1,0,1)$, and weights $(0.4,0.6)$. The coefficients of the
target score are drawn with seed 271828. Bases of the context spaces are
normalized by their visitation probabilities, and the normalization of
every kernel is verified by summing over all output states.

The sample experiment on the grid uses couplings $0.04$ and $0.08$, sample
streams with seeds 11, 22 and 33, and nested sample sizes
$1024,4096,16384,65536$ and $262144$. The graph and its weights are not
given to the estimator. The uncorrected comparison uses the same samples
and pair moments without inverse retention, and the exact oracle is
evaluated separately. The complete curve at $\epsilon=0.04$ is:
\begin{center}
\begin{tabular}{rrrr}\toprule
Samples & Uncorrected moments & Inverse retention & Oracle mixture\\\midrule
1,024&0.010350&0.015506&0.003022\\
4,096&0.004650&0.003294&0.003022\\
16,384&0.003558&0.001034&0.003022\\
65,536&0.003159&0.000303&0.003022\\
262,144&0.003068&0.000148&0.003022\\\bottomrule
\end{tabular}
\end{center}
These runs evaluate the unregularized estimator; for the regularized
family~\eqref{eq:estimated}, the population tradeoff is checked by direct
linear solves. The weighted design check uses the ten vertex pairs of five
vertices in lexicographic order, with weights
$(1,0.3,0.7,0.2,1.2,\allowbreak0.4,0.8,0.9,0.25,0.6)$, and includes all nontrivial
cuts with vertex zero on the first side. SLSQP minimizes the convex
criterion with analytic gradients, and we report the first-order optimality
gap on the simplex after normalizing the returned weights. All operator,
graph, attenuation and sample-estimation calculations use full-state sums
in float64.

\section{Further remarks on finite mixtures}
\label{S-app:remarks}
\paragraph{Finite-mixture bounds for a growing number of draws.}
The inequalities~\eqref{eq:finiteM-monotonic} concern expectations over
schedules; a realized Monte Carlo NLL need not decrease as further
independently drawn schedules are added. The identity
$\calL_1=\calL_{\rm path}$ uses the specified reveal law and its visitation
weights, and a random-mask loss with different weights can differ from it.

\paragraph{Posterior weights of the schedules.}
For a uniform library with $K$ components, define the posterior weights
(responsibilities) $r_s(x)=q_s(x)/\sum_uq_u(x)$ and the effective sample
size $\mathrm{ESS}(x)=1/\sum_s r_s(x)^2$. At the product reference all
responsibilities equal $1/K$. If $u_s$ are the component density scores,
then for each fixed state
\[
 \frac{\mathrm{ESS}(x)}K
 =1-\epsilon^2\operatorname{Var}_s[u_s(x)]+o(\epsilon^2).
\]
Thus the responsibilities are spread out in the local limit, but the theory
gives no universal threshold on the ESS at finite $\epsilon$. A low ESS
indicates a concentrated posterior over schedules and does not affect the
exact summation of a finite library. If one component dominates, its
likelihood also pays the normalization of the mixture: exactly,
$-\log q_\pi(x)=\min_s[-\log q_s(x)]+\log K+\log r_{\max}(x)$.

\end{document}